\documentclass[times]{elsarticle}
\usepackage[margin=1in]{geometry}
\usepackage{hyperref}

\usepackage{graphicx}
\usepackage{amssymb}

\usepackage{float}

\usepackage[subrefformat=parens]{subcaption}
\usepackage{amsmath}
\usepackage{bm}

\biboptions{sort&compress}

\journal{Journal of Molecular Graphics and Modeling}

\begin{document}

\begin{frontmatter}


\title{Visualizing Convolutional Neural Network Protein-Ligand Scoring}



\author[pitt]{Joshua Hochuli}
\author[pitt]{Alec Helbling}
\author[pitt]{Tamar Skaist}
\author[pitt]{Matthew Ragoza}
\author[pitt]{David Ryan Koes\corref{cor1}}
\ead{dkoes@pitt.edu}
\cortext[cor1]{Corresponding author}

\address[pitt]{Department of Computational and Systems Biology, University of Pittsburgh, 3501 Fifth Ave., Pittsburgh, PA 15260, United States}

\begin{abstract}
Protein-ligand scoring is an important step in a structure-based drug design pipeline. Selecting a correct binding pose and predicting the binding affinity of a protein-ligand complex enables effective virtual screening. Machine learning techniques can make use of the increasing amounts of structural data that are becoming publicly available. Convolutional neural network (CNN) scoring functions in particular have shown promise in pose selection and affinity prediction for protein-ligand complexes.

Neural networks are known for being difficult to interpret. Understanding the decisions of a particular network can help tune parameters and training data to maximize performance.  Visualization of neural networks helps decompose complex scoring functions into pictures that are more easily parsed by humans. Here we present three methods for visualizing how individual protein-ligand complexes are interpreted by 3D convolutional neural networks. We also present a visualization of the convolutional filters and their weights. We describe how the intuition provided by these visualizations aids in network design. 
\end{abstract}


\end{frontmatter}


\section{Introduction \label{intro}}

Protein-ligand scoring is an important computational method in a drug design pipeline \cite{Warren2006, Kitchen2004, Wang2003, Cheng2009, Cheng2012, csar2010}. 
In structure-based drug design methods, such as molecular docking, scoring is an essential subroutine that distinguishes among correct and incorrect binding modes and ranks the probability that a candidate molecule is active.  Improved scoring methods will result in more effective virtual screens that more accurately identify enriched subsets of drug candidates, providing more opportunities for success in successive stages of the drug discovery pipeline.

The wealth of protein-ligand structural and affinity data enables the development of scoring functions based on machine learning \cite{Ashtawy2015,jorissen2005,Ballester2010,chupakhin2013predicting,zilian2013sfcscore,durrant2011nnscore,sminapaper,brenner2016predicting}.  
Of particular interest are methods that use convolutional neural networks (CNNs) \cite{Ragoza2017, wallach2015atomnet,duvenaud2015convolutional,schutt2017moleculenet,gomes2017atomic,kdeep} to recognize potent protein-ligand interactions, as CNNs have been remarkably successful at the analogous image recognition problem \cite{krizhevsky2012imagenet,szegedy2015going,msdeepresidual}.
Unlike force field or empirical scoring functions, whose functional form is designed to represent known physical interactions such as hydrogen bonding or steric interactions, machine learning methods can derive both their model structure and parameters directly from the data.  However, this increase in model expressiveness comes at the cost of reduced model interpretability.

The lack of interpretability of a CNN model presents challenges both when developing a scoring function and in understanding its application. Choosing input representations, managing training and test data, and determining optimal parameters all depend on understanding how the CNN behaves. Simple ``black box'' treatment of the model is not sufficient to guide such decisions.  Additionally, visualizations can provide human-interpretable insights to help guide medicinal chemistry optimization.

In the image classification domain, there are number of methods that provide insight into the inner workings of of a trained CNN by projecting network decisions back on the readily visualized input image.  These methods reveal what parts of an input image are important \cite{yosinski2015,samek2017} and how the input is represented at different layers in the network \cite{mahendran2016}. Loss gradients have also been visualized in order to determine what parts of an input contribute most to an incorrect prediction \cite{simonyan2013}. Here we investigate grid-based CNN scoring of protein-ligand complexes and show how network decisions can be projected back to an atomistic granularity.

We visualize the convolutional filters of the first layer of the network to gain insight in the initial featurization learned.  In order to gain atomistic insight into specific network decisions (e.g., why a ligand is scored as having a high/low affinity), we introduce and compare three methods for projecting the network's decision onto the molecular input: masking, gradient, and conserved layer-wise relevance propagation (CLRP).
CLRP is a novel refinement of layer-wise relevance propagation (LRP) \cite{bach-plos15,lapuschkin-jmlr16} that better compensates for zero-weight activations.  This is important since such activations emerge naturally from ``empty'' space in the input where there are no protein or ligand atoms (e.g. implicit solvent).  This enables visualizations that account for the contributions of solvent to the final prediction of the network.
 
We apply each method to a network that was trained for both pose selection (distinguish low-RMSD from high-RMSD poses) and affinity prediction. Convolutional filter visualization provides insight into the low-level features identified by the network.  We compare and contrast the three atomistic visualizations and show how they provide different insights and have different properties.  Our visualization implementations and CNN models are available under an open-source license as part of gnina, our framework for structure-based deep learning based off of AutoDock Vina \cite{vina} and Caffe \cite{jia2014caffe}, at \url{https://github.com/gnina}. 

\section{Methods \label{methods}}

\begin{figure}[tbp]
\centering
\centerline{\includegraphics[width=\linewidth]{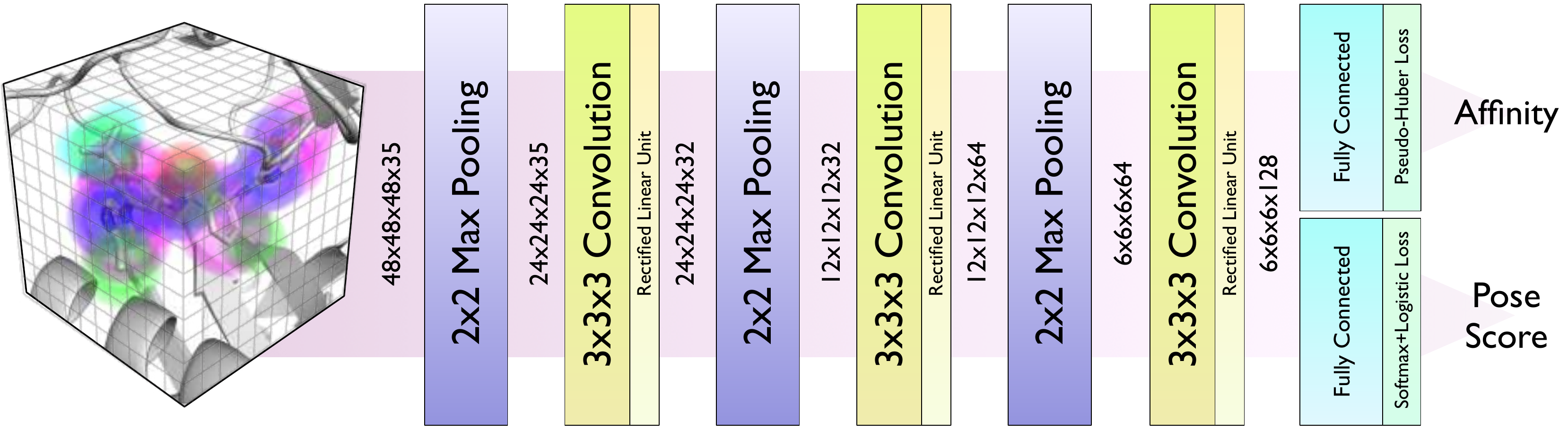}}
\caption{\label{fig:architecture}Architecture of the network used to evaluate visualization methods.  The input is a voxelized grid of Gaussian atom type densities.}
\end{figure}

After describing the design and training of a CNN model for pose scoring and affinity prediction,
we describe an approach for analyzing the learned weights of the first layer of a grid-based CNN model and three distinct methods for mapping a CNN prediction back onto the atomic input.  

\subsection{Training}
\begin{figure}[tbp]
\centering
\begin{subfigure}[t]{.48\linewidth}
\includegraphics[width=\linewidth]{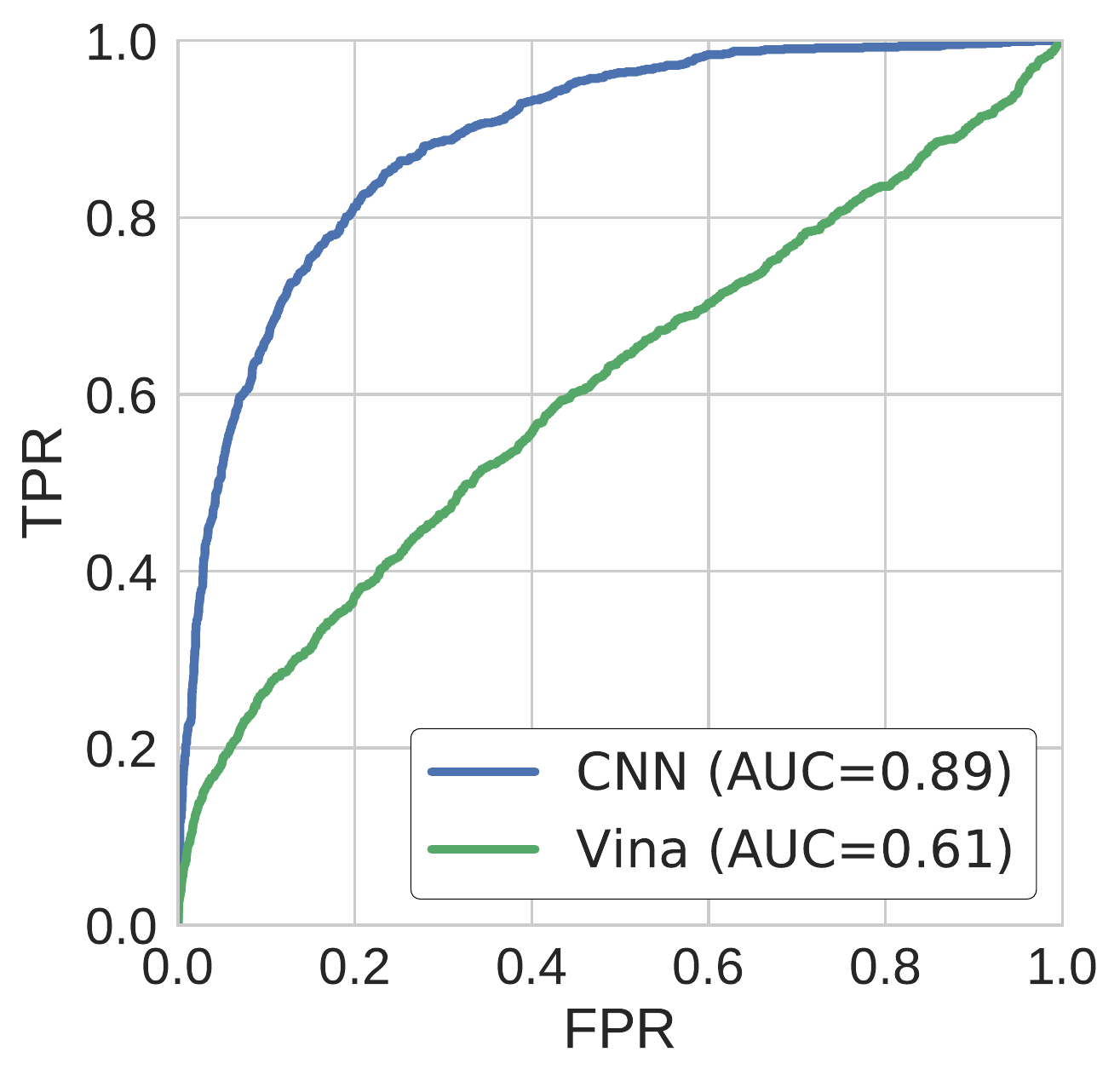}
\caption[]{\label{poseeval}}
\end{subfigure}%
\hfill
\begin{subfigure}[t]{.48\linewidth}
\includegraphics[width=\linewidth]{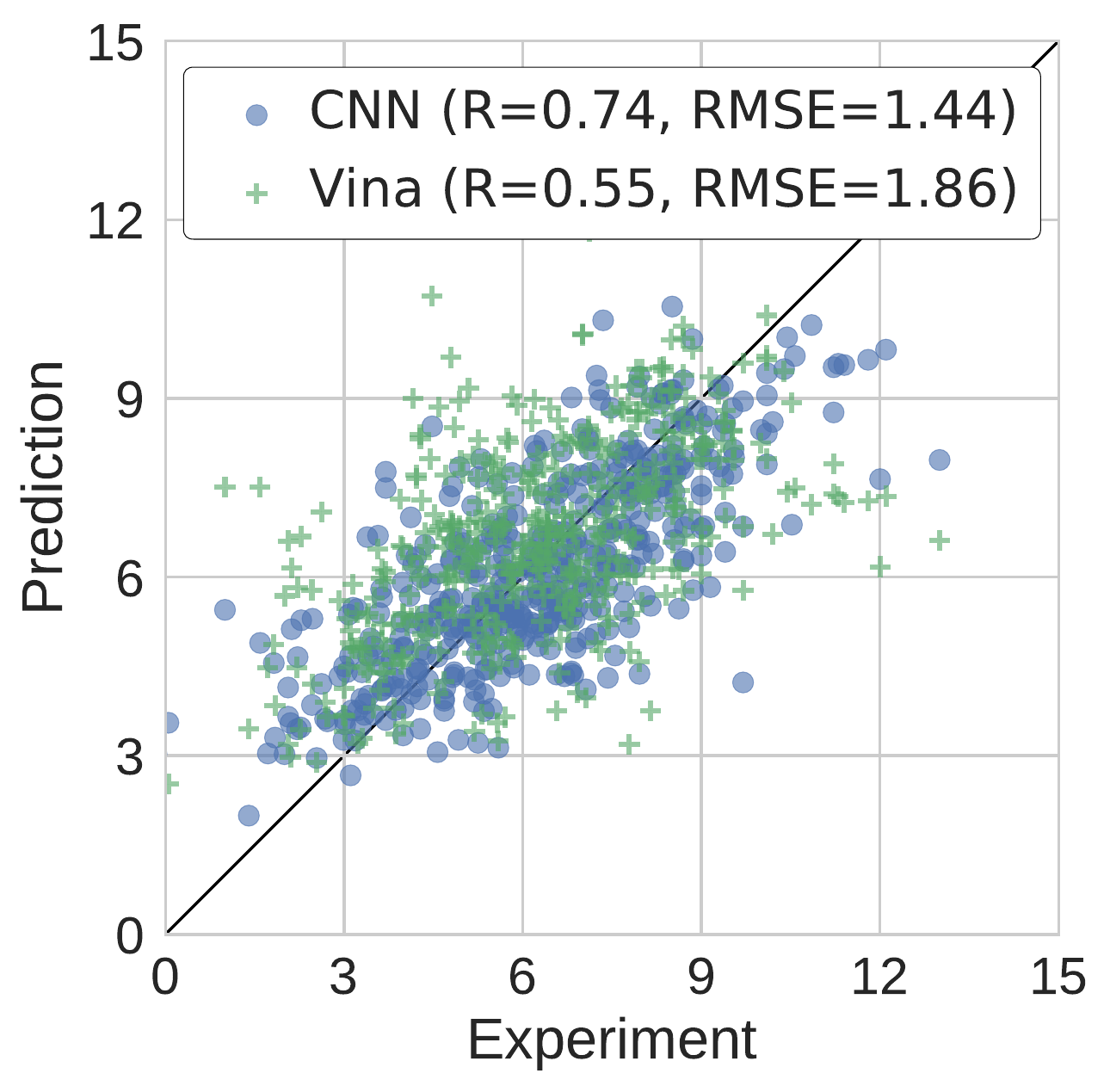}
\caption[]{\label{affeval}}
\end{subfigure}\caption{\label{fig:eval} CNN model performance. \subref{poseeval} Discriminative ability of CNN scoring and AutoDock Vina scoring to distinguish between low ($< 2${\AA} RMSD) and high ($> 4${\AA} RMSD) docked poses across the entire CSAR evaluation set. The AUC, area under the ROC curve, which plots the true positive rate (TPR) with respect to the false positive rate (FPR) as the classification threshold is increased, is substantially better for the CNN model. \subref{affeval} Affinity prediction performance.  The best score of all generated poses for each target is used.  Values are log affinity units, as provided in the CSAR data set. Vina energies are Boltzmann scaled appropriately ($T=298K$).
}
\end{figure}

\begin{table}
\begin{minipage}[t]{0.4\linewidth}
\begin{tabular}[t]{c}
Receptor Atom Types \\ \hline
AliphaticCarbonXSHydrophobe \\
AliphaticCarbonXSNonHydrophobe \\
AromaticCarbonXSHydrophobe \\
AromaticCarbonXSNonHydrophobe \\
Calcium \\
Iron \\
Magnesium \\
Nitrogen \\
NitrogenXSAcceptor \\
NitrogenXSDonor \\
NitrogenXSDonorAcceptor \\
OxygenXSAcceptor \\
OxygenXSDonorAcceptor \\
Phosphorus \\
Sulfur \\
Zinc
\end{tabular}
\end{minipage}
\hfill
\begin{minipage}[t]{0.4\linewidth}
\begin{tabular}[t]{c}
Ligand Atom Types \\ \hline
AliphaticCarbonXSHydrophobe \\
AliphaticCarbonXSNonHydrophobe \\
AromaticCarbonXSHydrophobe \\
AromaticCarbonXSNonHydrophobe \\
Bromine \\
Chlorine \\
Fluorine \\
Nitrogen \\
NitrogenXSAcceptor \\
NitrogenXSDonor \\
NitrogenXSDonorAcceptor \\
Oxygen \\
OxygenXSAcceptor \\
OxygenXSDonorAcceptor \\
Phosphorus \\
Sulfur \\
SulfurAcceptor \\
Iodine \\
Boron 
\end{tabular}
\end{minipage}
\caption{\label{atomtypes} The 35 atom types used in gnina.  Carbon atoms are distinguished by aromaticity and adjacency to polar atoms (``NonHydrophobe'').  Polar atoms are distinguished by hydrogen bonding propensity. }
\end{table}

For our CNN model, we extend our previously described architecture \cite{Ragoza2017} as shown in Figure~\ref{fig:architecture}.  The atoms of the input complex are represented using truncated Gaussians and 35 distinct atom types, shown in Table~\ref{atomtypes}.  This continuous representation is discretized onto a cubic grid that is 24{\AA} on each side and has a resolution of 0.5{\AA}.  The input is fed through three units of max pooling and convolution with rectified linear unit (ReLU) activation functions after which the result is split into two separate fully connected layers with no hidden units. One fully connected layer is trained to score poses and generate a probability that an input pose is a low ($<2${\AA}) RMSD pose using a softmax layer (which scales predictions to be between zero and one) and a logistic loss function.  The other is trained to predict the binding affinity in log units using a pseudo-Huber loss function. This loss interpolates between an L2 and L1 loss according to a parameter $\delta$ to reduce outlier bias:
\begin{eqnarray}
L(y,\hat{y}) = \delta^2 \sqrt{1+\left(\frac{y-\hat{y}}{\delta}\right)^2} - \delta^2
\end{eqnarray}
As the training set includes incorrect ($>4${\AA} RMSD) poses, for which the
correct binding affinity is not well-defined, a hinge loss is used so that a loss is only incurred if the affinity is predicted to be too high.  The complete model used for training is available at \url{https://github.com/gnina/models}.

For training data we use a set of poses generated by redocking the ligands of the the 2016 PDBbind refined set \cite{pdbbind2016}. Poses were generated using the AutoDock Vina scoring function \cite{vina} as implemented in smina \cite{sminapaper}. The binding site for docking was defined using the pocket residues specified in the PDBbind.  The input ligand conformation was generated from 2D SMILES using RDKit \cite{rdkit}.
To increase the number of low RMSD poses in the training set, the docked poses
were supplemented by energy minimized crystal poses.  To avoid introducing
training artifacts related to the creation of crystal structures (e.g. bond
lengths or angles unique to crystal poses), these crystal ligands were optimized
independently of the receptor using the UFF forcefield of RDKit to replicate
the conformer generation process and then minimized with respect to the
receptor using smina.  The training set was then further expanded by three
rounds of iteratively training a model, using the trained model to refine the
docked poses \cite{ragoza2017ligand}, and adding these refined poses to the training set.  This
iterative processes extends the training set to include poses and conformations
that are not biased to the Vina energy potential. The final training set
contains 255,035 ligand-receptor complexes, of which 15,814 are less than 2{\AA} RMSD from the crystal pose.

Using this training set, we trained our model for 150,000 iterations with a batch size of 50 using our customized version of the GPU-optimized Caffe deep learning framework \cite{jia2014caffe}.  Each batch was balanced to contain an equal number of positive and negative examples (low and high RMSD poses) as well as stratified by receptor so that every receptor target was uniformly sampled, regardless of the number of docked structures.  At each iteration, a random rotation and translation is applied to every input complex in order to prevent the network from learning coordinate-frame dependent features.

The performance of the trained model on docked poses created from the
high-quality and compact CSAR set \cite{csar2010} is shown in
Figure~\ref{fig:eval}.  The docked poses of the CSAR set were generated using
the binding site defined by the cognate ligand and smina. CNN scoring performs
substantially better than Vina scoring at distinguishing between low and high
RMSD poses, with an area under the ROC curve of 0.89.  CNN scoring also exhibits
better correlation with binding affinity, with an Pearson correlation
coefficient of 0.74. This is comparable to the best performing scoring functions
in the original CSAR evaluation \cite{csar2010}, although our evaluation is
performed on an expanded benchmark and uses docked, instead of crystal, poses.
We emphasize our purpose in this evaluation is \textit{not} to robustly measure generalization error; 60\% of the CSAR complexes are duplicated in the PDBbind refined training set and this results in an overly optimistic assessment of general model performance.  Instead, our focus here is to demonstrate that the trained model generates meaningful results for this test set, as this is a prerequisite for achieving our primary goal of meaningful visualizations.

\subsection{Convolutional Filter Visualization}
The convolutional layer of a CNN consists of filters that are applied to the input to produce feature maps that represent the presence of a specific learned local feature of the input. As the filters are linear functions of the input, the weights, $W_f$ of the filter and an additive bias, $b_f$, dictate the contribution of the input, $x$, to the activation function, $\sigma$, that generates the output:
\begin{eqnarray}
\mathrm{output}_f = \sigma(W_f x + b_f)
\end{eqnarray}
These weights are only amenable to interpretation in the first convolutional layer, where the input corresponds directly to atom type densities.  In our network (Figure~\ref{fig:architecture}), there are 32 convolutional filters in the first layer, each consisting of 35x3x3x3 weights corresponding to 35 atom types and a 3x3x3 volume.  Analysis of these initial filters provides insight into how the network interprets different atom types and demonstrates that the initial layer of the network is conditioned to identify certain patterns.

\subsection{Masking}

\begin{figure}[tbp]
\centering
\centerline{\includegraphics[width=0.8\linewidth]{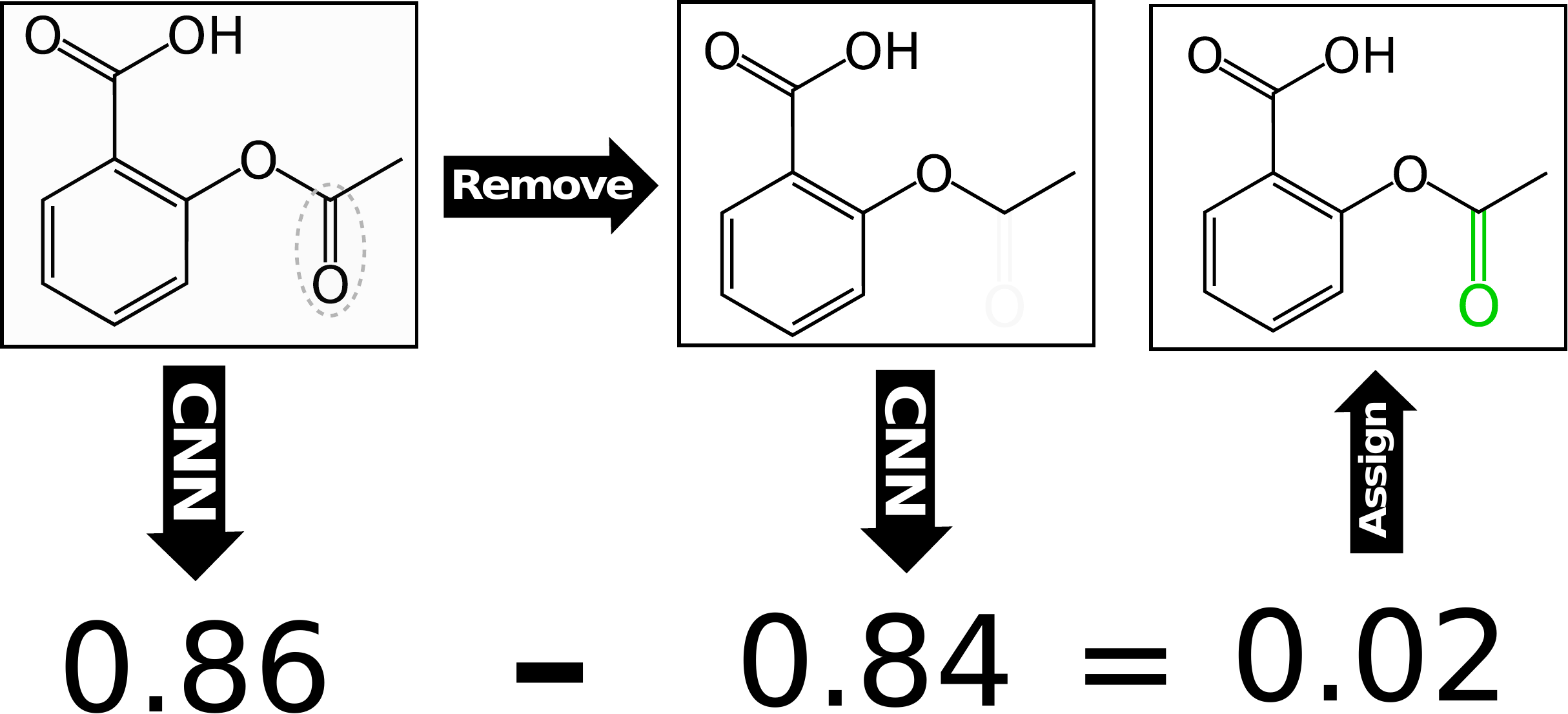}}
\caption{\label{fig:difference}Simplified visual of the masking algorithm }
\end{figure}

Masking evaluates the sensitivity of the model to changes in the input by removing (masking) part of the input and computing the difference in the predicted output score between the original and masked input \cite{szegedy2013NIPS}. For our molecular inputs, this is accomplished for the ligand both by removing individual atoms, as illustrated in Figure~\ref{fig:difference} and molecular fragments. 
Fragments are generated with the RDKit \cite{rdkit} cheminformatics package by
enumerating all heavy-atom subgraphs of the ligand containing up to six bonds.
The score difference with the fragment removed is evenly distributed among the constituent atoms of the fragment.  These fragment atomic averages are summed to compute a fragment masking value for each atom.  For both atom and fragment removals, atom types are not recomputed (e.g., removing an atom from an aromatic ring will not cause the remaining atoms to be represented as aliphatic).  Individual atom masking results in a granular assessment of atomic contributions to the final score, while fragment masking accounts for the contributions of entire functional groups and, by averaging across groups of atoms, results in a smoother distribution of atomic contributions.  To get the best of both approaches, our masking visualization uses the average of individual atom masking and fragment masking.

Masking-based visualization of the receptor is performed by removing entire residues and distributing the score difference evenly among the atoms of the residue. A residue granularity is used since the number of receptor atoms in a typical binding site makes finer-grained masking computationally demanding and because residues are natural building blocks for analyzing contributions to binding in proteins. 

Masking is computationally demanding because the number of neural network evaluations needed grows linearly in the size of the receptor and polynomially in the size of the ligand (due to the fragment generation). In contrast, the gradient and LRP approaches that follow require only a single backwards pass through the neural network.

\subsection{Atomic Gradient}

Neural network training typically involves gradient-based optimization in order to minimize a loss function, which necessitates the computation of the loss gradient on the network parameters--that is, the partial derivatives of the loss function with respect to those parameters--through backpropagation. This method can be extended to compute the gradient on the network input as well, allowing visualization of the gradient as three-dimensional vectors on each atom. This gradient on atom coordinates can give insight into how the input should be changed in order to more fully realize the desired class label (e.g., a low-RMSD pose). To calculate the gradients, a forward pass is performed, the loss is calculated with respect to the desired class label (e.g. a true label for pose classification), and  a backward pass then computes the loss gradient on atomic coordinates. The negative of this  gradient can be visualized as the direction in atomic coordinate space to move the atom to minimize the loss for the class, increasing the probability that the network classifies the pose as a true binding pose. This approach works just as well with an affinity prediction CNN, in which case the atomic gradient is computed to indicate the direction that increases the network's predicted binding affinity.

We designed our CNN scoring function to be fully differentiable with respect to atomic coordinates by using a custom atomic grid input representation with an analytic derivative. Formally, the CNN is a function $f$ that takes as input atomic coordinates and atom types and maps to an output value that can be a probability distribution on class labels or a real-valued affinity prediction. Atom coordinates are discretized to a four-dimensional grid, $\textbf{G}$, that is a vector of three-dimensional grids of atom density (channels), one for each atom type. The density of a particular atom at a grid point $g$ in the grid channel corresponding to its atom type is given as a function of the atom's Van der Waals radius $r$ and the distance $d$ between the atom and the grid point:

\begin{equation}
g(d, r) =
\begin{cases}
e^{-\frac{2{d}^2}{{r}^2}} & 0 \leq d < r \\
\frac{4}{e^2r^2}{d}^2 - \frac{12}{e^2r}d + \frac{9}{e^2} & r \leq d < 1.5r \\
0 & d \geq 1.5r \\
\end{cases} \\
\end{equation}

This function is differentiable with respect to distance as follows:

\begin{equation}
\frac{\partial g}{\partial d} =
\begin{cases}
-\frac{4d}{r^2}e^{\frac{-2{d}^2}{{r}^2}} & 0 \leq d \leq r \\
\frac{8}{e^2r^2}d - \frac{12}{e^2r} & r < d < 1.5r \\
0 & d \geq 1.5r \\
\end{cases} \\
\end{equation}

The gradient of the scoring function with respect to atom coordinates $a$, $\frac{\partial f}{\partial \textbf{a}}$, is computed by applying the chain rule and summing over the grid points with the appropriate atom type that overlap the atom, ${G}_a$:

\begin{equation}
\frac{\partial f}{\partial \bm{a}} = \sum_{g \in \bm{G}_a} \frac{\partial f}{\partial g} \frac{\partial g}{\partial d} \frac{\partial d}{\partial \bm{a}} \\
\end{equation}

The resulting gradient can then be visualized either as vectors in 3D space or as scalar magnitudes on each atom. The atom gradient approach provides qualitatively different information than either masking or LRP. Since it indicates how changes in spatial relationships affect the output score, this can help to understand how the network wants to modify the input. However, this method does not try to explain how the network arrived at the current score by assigning relative importance to different components. For example, a locally optimal ligand (from the perspective of the network) would have all zero gradients, which provides no insight into what makes the ligand desirable.

\subsection{CLRP}

Layer-wise relevance propagation (LRP) \cite{bach-plos15} maps the output of a neural network, such as a classification probability, back to the original input. It introduces a
quantity called ``relevance'' that is initialized as the network output. The relevance is then propagated back through the network until it
reaches the input. The propagation is performed proportionally to the input activations ($z_{ij} = x_iw_{ij}$) of each layer, such that the relevance of node $i$ in layer $l$ is the sum of the relevances of its successor nodes, $j$, weighted by the activation value generated along the edge $z_{ij}$ during the forward pass:

\begin{equation}
R_i^{(l)} = \sum_j \frac{z_{ij}}{\sum_{i'j}}R_j^{(l+1)}
\end{equation}

This construction maintains the invariance that the relevance at
each layer is conserved; the sum of the relevance assigned to each node $d$, $R_d$, at each layer $l$ is
exactly the same, so the classification decision is distributed across the
input by the end of the backward pass:

\begin{equation} \label{conservation}
f(x) = \ldots = \sum_{d \in l+1}{R_{d}^{(l+1)}} = \sum_{d \in l}{R_{d}^{(l)}} = \ldots = \sum_{d}{R_{d}^{(1)}}
\end{equation}

Like the gradient approach, this method also has the advantage of
requiring a single backward pass through the network in order to visualize an
example, rather than hundreds or thousands of forward passes in the case of the masking algorithm.  Unlike the gradient approach, LRP is not distributing the gradient of the loss throughout the network; instead it is distributing the output value itself as an explanation for why a particular input generated that value.

Issues arise with LRP when attempting to propagate through nodes with zero
activations. Using the basic algorithm, the relevance propagated backwards through
those nodes becomes unbounded, violating the relevance conservation. Previously, two methods have been proposed to compensate for these zero-activation ``dead'' nodes \citep{bach-plos15}. The first, alpha-beta decomposition, is to separate negative and positive pre-activation values, and use the proportions of
the negative and positive values together to propagate relevance. The second is
to introduce a stabilizing factor $\epsilon$ which draws ``dead'' nodes away from
values close to zero. Both approaches violate the conservation property (\ref{conservation}).

We propose a third method for conserving relevance in the presence of dead
nodes, \textit{conserved layer-wise relevance propagation} (CLRP). Rather
than attempt to work around dead nodes by either attenuating them (in the case
of epsilon stabilization) or artificially weighting positive or negative
pre-activations (alpha-beta decomposition), relevance directed
onto dead nodes is instead redistributed proportionally across the remaining
nodes in the layer. This results in dead nodes passing no relevance backward, as they pass no pre-activations in the forward pass. The quantity $S_{l}$ is the total relevance belonging to nodes whose pre-activation $z_{j} = 0$ at layer $l$:

\begin{equation} \label{Sl}
S_{l}= \sum_{j}{
\begin{cases}
      0 & z_{j} \neq 0 \\
      R_{j} & z_{j} = 0
\end{cases}}
\end{equation}

After computing $S_{l}$, the adjusted relevance for node $j$ can be calculated as follows, where $Z_{l}$ is the sum of pre-activations across layer $l$:

\begin{equation}
R_{j} = 
\begin{cases}
      0 & z_{j} = 0 \\
      R_{j} + \frac{z_{j}}{Z_{l}}*S_{l} & z_{j} \neq 0
\end{cases}
\end{equation}

The robustness of CLRP is necessary to meaningfully apply LRP to our discretized molecular inputs, since significant portions of the input may have zero atom density values, corresponding to implicit solvent.  These empty areas generate ``dead'' nodes in the first layer.  In addition to compensating for these nodes with CLRP, we retain their contributions to $S_l$ (\ref{Sl}). In the first convolution layer these values directly map to input Cartesian coordinates and we can use them to visualize the contributions of empty space (implicit solvent) to the final score.

\section{Results \label{results}}

We first present visualizations of the CNN filter weights, which provide insights into what low-level features the network is learning, and then provide examples of our three different atomic visualizations:  masking, gradients, and CLRP.  These three visualizations are then systematically compared.  Finally we demonstrate the unique ability of CLRP to provide visualizations of the importance of empty space.

\begin{figure}[tbp]
\centering
\includegraphics[width=\linewidth]{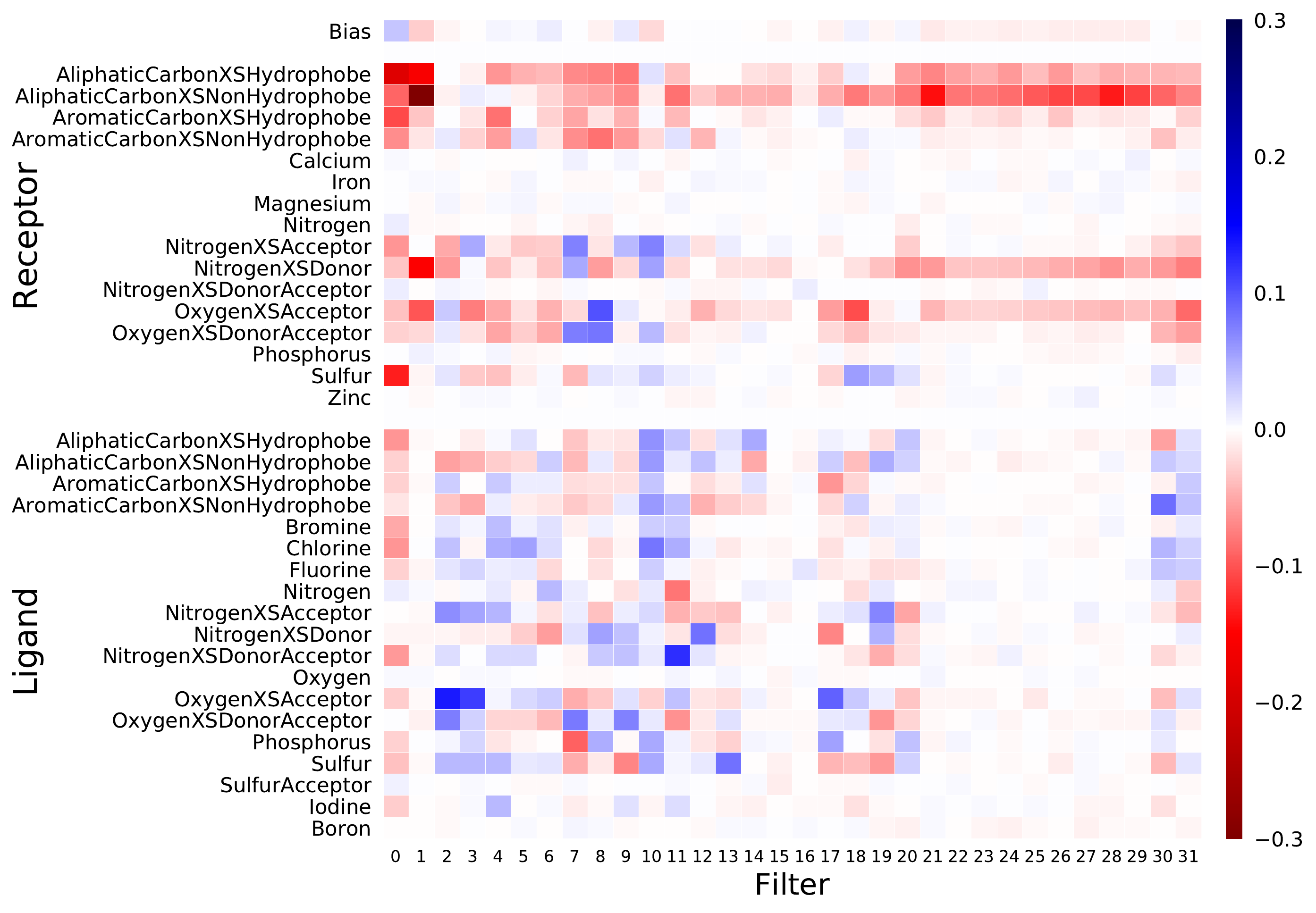}
\caption{\label{fig:avefilters}  The average weight within each atom type channel of the first 32 convolutional filters of the CNN.  Filters are shown clustered by Euclidean distance so that similar filters are grouped together.  The bias is divided by 27 to match the scaling of the weight averages.}
\end{figure}

\begin{figure}[tbp]
\centering
\includegraphics[width=\linewidth]{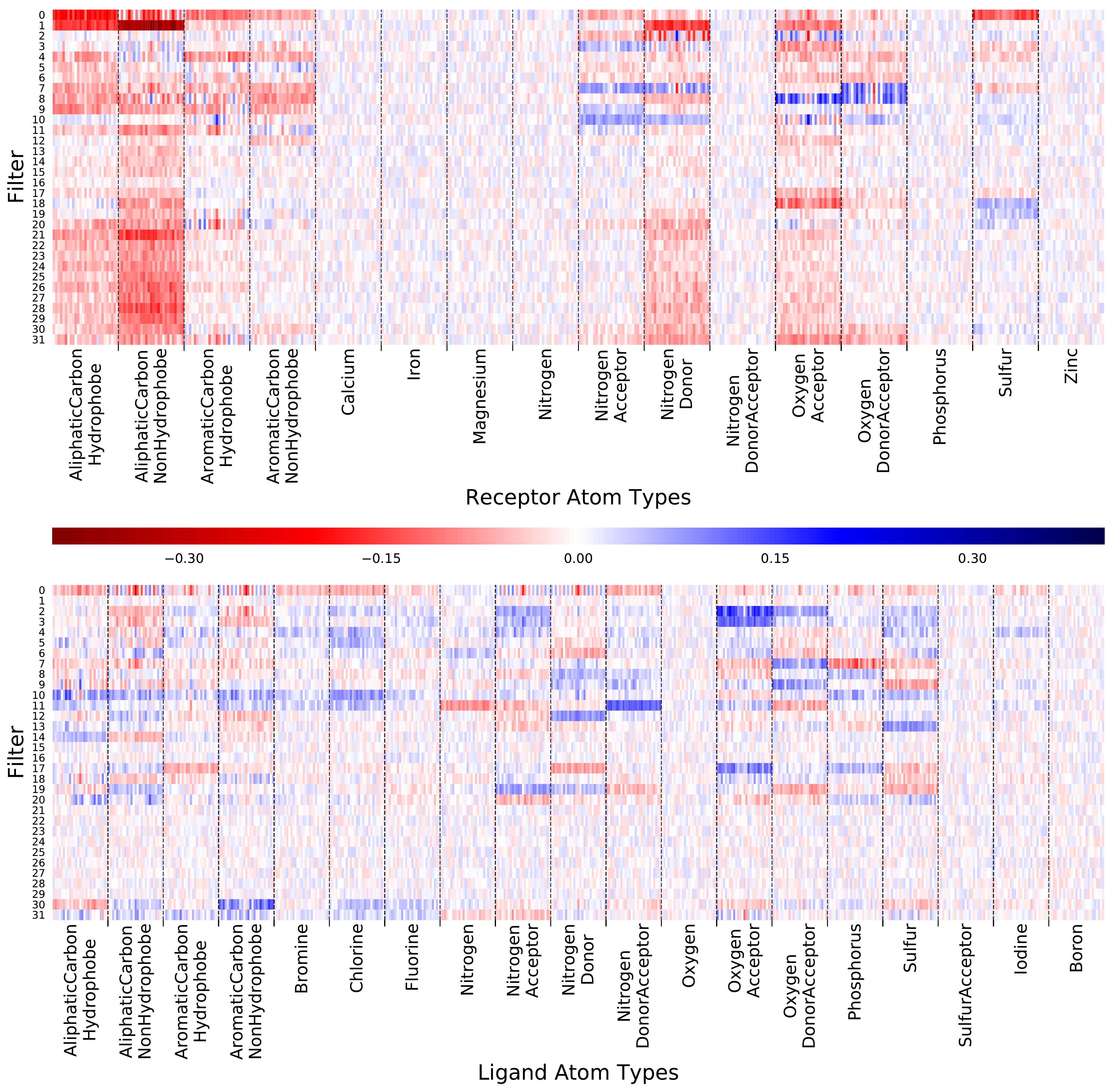}
\caption{\label{fig:flatfilters} The full weight vector of each filter, segmented by atom type.  Within each atom type, the 27 weights of the 3x3x3 filter are shown flattened to a single dimension such that the center of the 1D array is the center of the 3D cube. Filters are shown in the same clustered ordering as in Figure~\ref{fig:avefilters}.}
\end{figure}

\subsection{Convolutional Filter Visualization}

The first layer of convolutional filters of the CNN reveal what low-level features of the input the CNN is recognizing for propagation to successive layers.  As the weights of these first filters are applied directly to the input, they can be interpreted in terms of atom densities and spatial relationships.  In our network, there are 32 filters in the first convolutional layer, each of which has 3x3x3x35 weights, corresponding to a uniform samples from a 3{\AA} cube across all 35 atom types.  These filters are visualized in terms of the average weight within each atom type channel in Figure~\ref{fig:avefilters}, that is, the average of the 3x3x3 weights that are applied to a single atom type.  This visualization shows what atom types maximally activate each filter, but spatial relationships are averaged out.  The average weight may be either positive or negative; the significance of the sign depends on the additive bias, also shown in Figure~\ref{fig:avefilters}.  Since the network uses ReLU activation functions, if applying the weights and bias to an input results in a negative number, the output of the filter will be zero.

Several informative patterns are present in Figure~\ref{fig:avefilters}.  Some filters focus on receptor types, others on ligand types, but most have a mix of positive and negative weights across multiple different atom types. Several atom types (e.g. metals) have low average weights across all filters.  These correspond to relatively rare atom types.  The lack of strong activations for rare atom types implies the network isn't overfitting to rare events and that it might be beneficial to merge these atom types into a single generic type.  While less precise, a generic type would reduce the dimensions of the input, resulting in faster classification.  There are nine filters (21 through 29) that have mostly negative weights, focused on common receptor atom types, and a negative bias.  Although it is possible these filters are recognizing subtle spatial features that are exclusive of the receptor, an alternative explanation is that these filters consistently generate a negative input to the activation function, resulting in a zero output.  That is, the network has learned to turn off these filters.  This suggests that reducing the number of initial filters may result an an equally accurate, but simpler, model.

The full weight vector of each filter is shown in Figure~\ref{fig:flatfilters}.  The spatial patterns of individual filters, partitioned by atom type, provide some insights into the structural features learned by the CNN.  Low-activation filters appear as washed-out noise. Within an atom type, some filters are largely uniform, merely registering the presence of an atom, while others have distinctive variations, both symmetric and asymmetric, that will respond differently to different spatial arrangements of atoms.  As an example, several of the hydrogen bonding atom types, both in the receptor in ligand, demonstrate banding patters where weights vary between positive and negative values.  A number of filters have a large central weight and surrounding weights of opposite signs, corresponding to the pattern of an edge detection filter.

Visualizing the initial convolutional filters provides insights into the low-level features recognized by the trained network.  However, since filter weights are intrinsic to the trained network itself and do not vary for different inputs, they do not provide insights as to why the network scored a given protein-ligand pose the way it did or how the compound could be modified to improve its score.  In the next sections, we describe multiple methods for projecting the neural network score back onto the molecular input at an atomistic granularity.

\begin{figure}[tbp]
\centering

\begin{subfigure}[t]{\linewidth}
    \begin{subfigure}[t]{.32\linewidth}
    \centering
    \includegraphics[width=\linewidth]{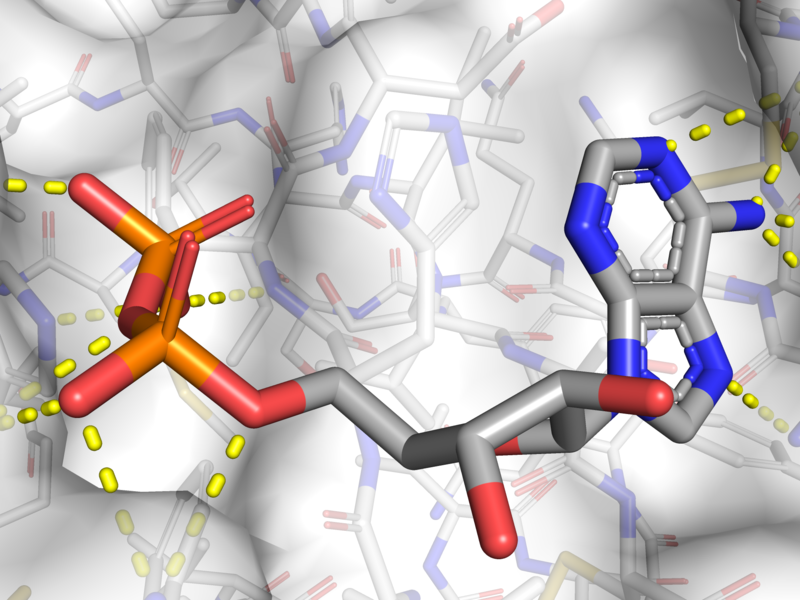}
    \caption{1o0h:  2.6977/0.2554}
    \end{subfigure}%
    \hfill
    \begin{subfigure}[t]{.32\linewidth}
    \centering
    \includegraphics[width=\linewidth]{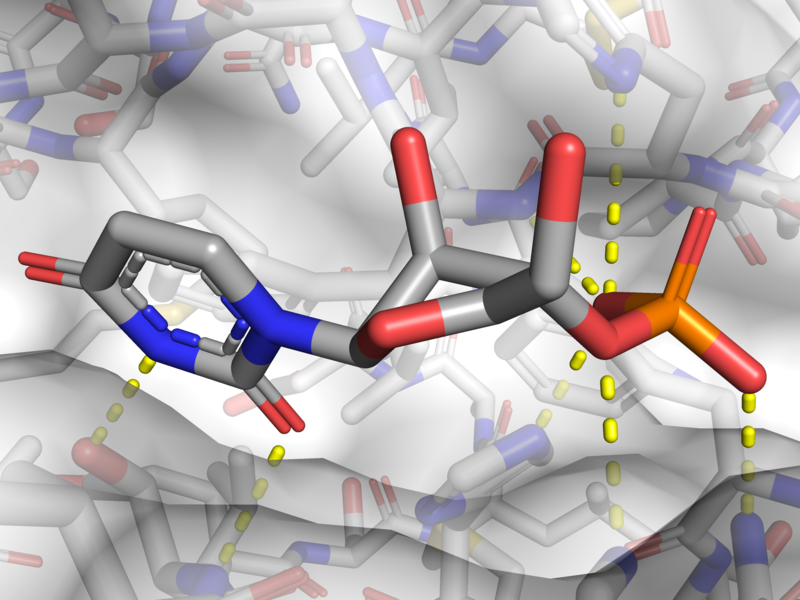}
    \caption{1w4o: 4.9329/0.9828}
    \end{subfigure}    
    \hfill
    \begin{subfigure}[t]{.32\linewidth}
    \centering
    \includegraphics[width=\linewidth]{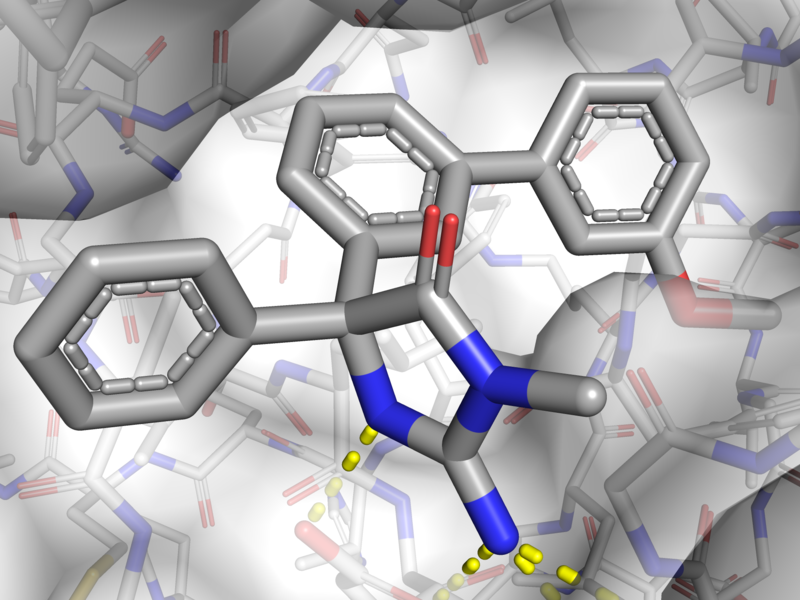}
    \caption{4djv: 5.9512/0.8941}
    \end{subfigure}
\end{subfigure}
\caption{Complexes colored by atom type shown with their affinity/pose score.}
    \label{fig:blank_examples}
\end{figure}

\begin{figure}[tbp]
\centering

\begin{subfigure}[t]{\linewidth}
    \caption*{Affinity Prediction Score = 2.6977}
    \begin{subfigure}[t]{.32\linewidth}
    \centering
    \includegraphics[width=\linewidth]{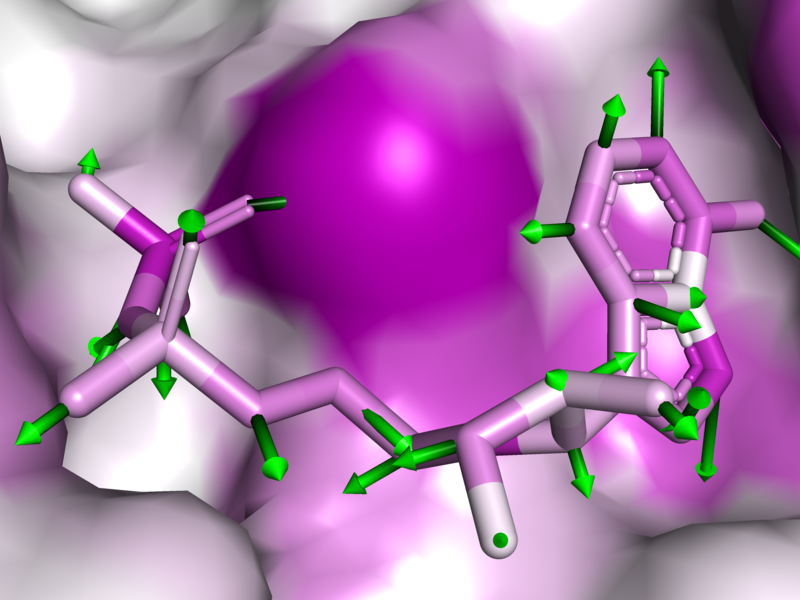}
    \caption{Gradient}
    \end{subfigure}%
    \hfill
    \begin{subfigure}[t]{.32\linewidth}
    \centering
    \includegraphics[width=\linewidth]{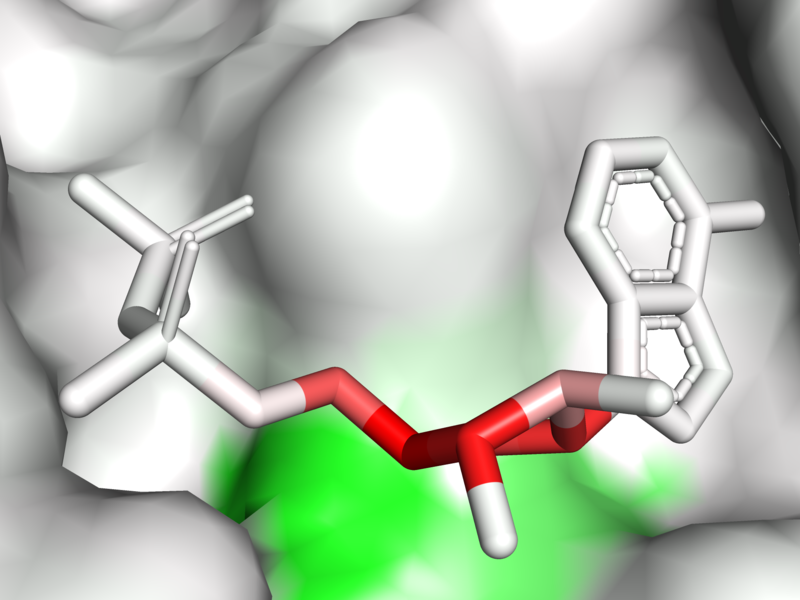}
    \caption{CLRP}
    \end{subfigure}
    \hfill
    \begin{subfigure}[t]{.32\linewidth}
    \centering
    \includegraphics[width=\linewidth]{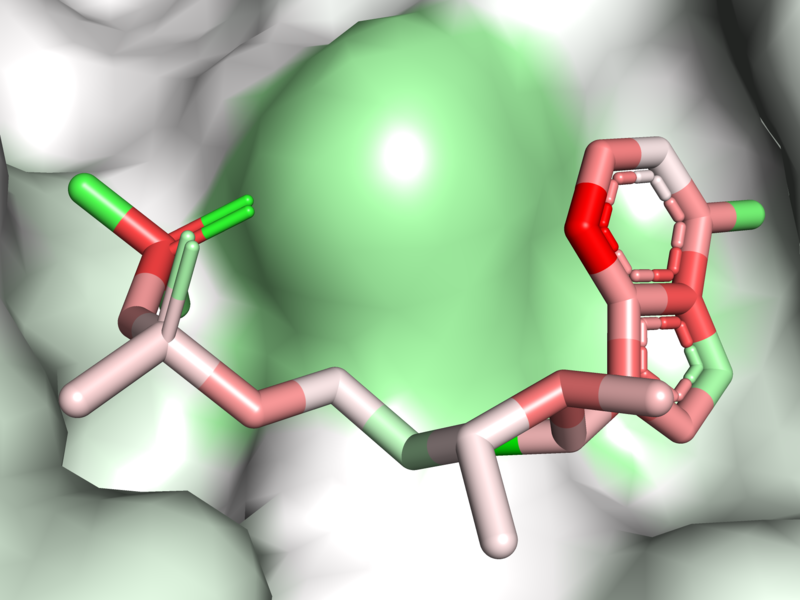}
    \caption{Masking}
    \end{subfigure}
\end{subfigure}

\begin{subfigure}[t]{\linewidth}
    \caption*{Pose Score = 0.2554}
    \begin{subfigure}[t]{.32\linewidth}
    \centering
    \includegraphics[width=\linewidth]{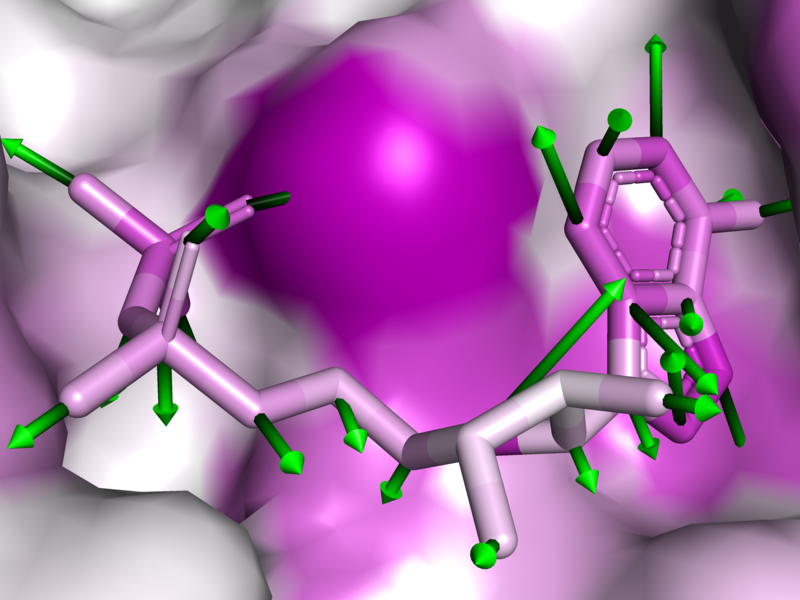}
    \caption{Gradient}
    \end{subfigure}%
    \hfill
    \begin{subfigure}[t]{.32\linewidth}
    \centering
    \includegraphics[width=\linewidth]{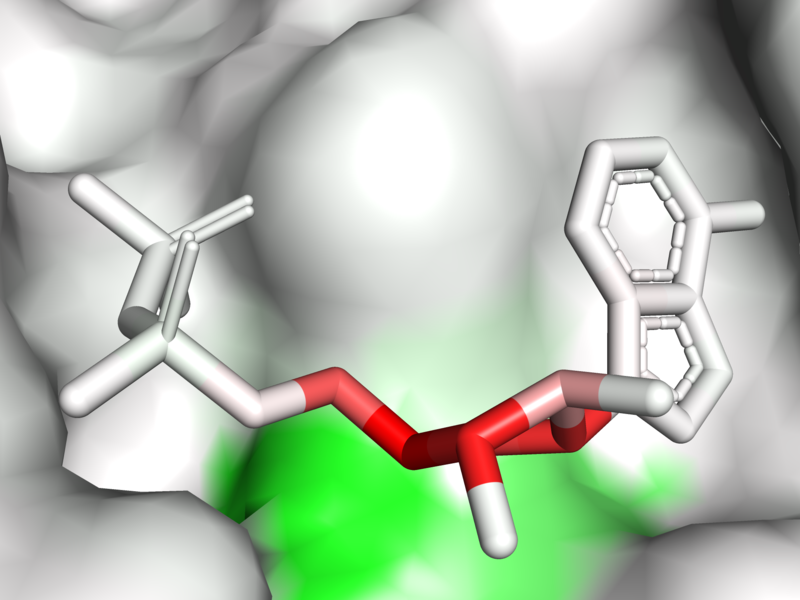}
    \caption{CLRP}
    \end{subfigure}
    \hfill
    \begin{subfigure}[t]{.32\linewidth}
    \centering
    \includegraphics[width=\linewidth]{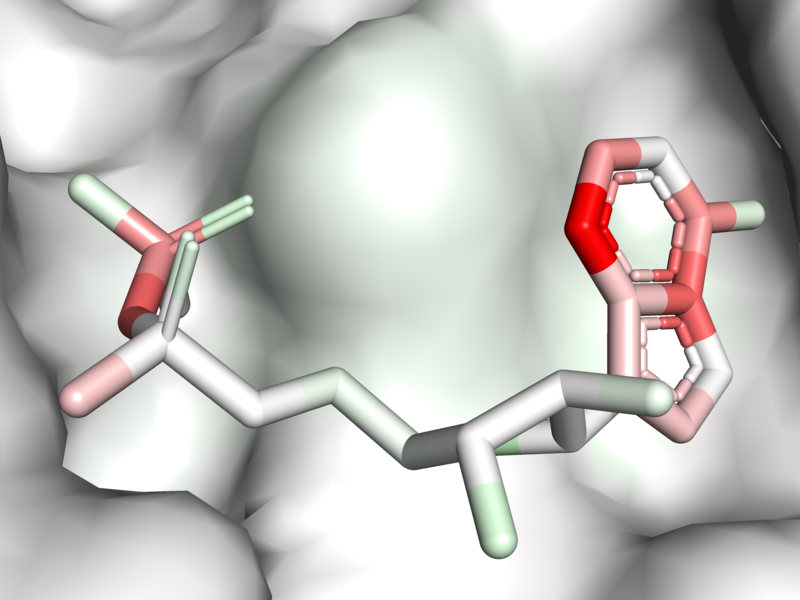}
    \caption{Masking}
    \end{subfigure}
\end{subfigure}
\caption{PDB 1o0h.  The atom scores calculated with the CLRP and masking algorithms are shown as a red-white-green gradient, with green corresponding to a more favorable score.  As the gradient norms are unsigned, they are visualized as a purple gradient.}
    \label{fig:1o0h}
\end{figure}

\begin{figure}[tbp]
\centering
\begin{subfigure}[t]{\linewidth}
    \caption*{Affinity Prediction Score = 4.9329}
    \begin{subfigure}[t]{.32\linewidth}
    \centering
    \includegraphics[width=\linewidth]{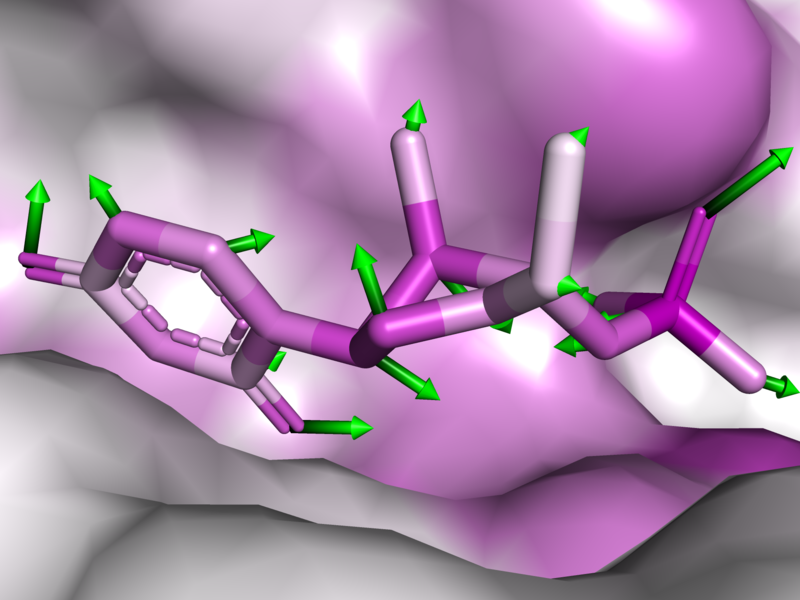}
    \caption{Gradient}
    \end{subfigure}%
    \hfill
    \begin{subfigure}[t]{.32\linewidth}
    \centering
    \includegraphics[width=\linewidth]{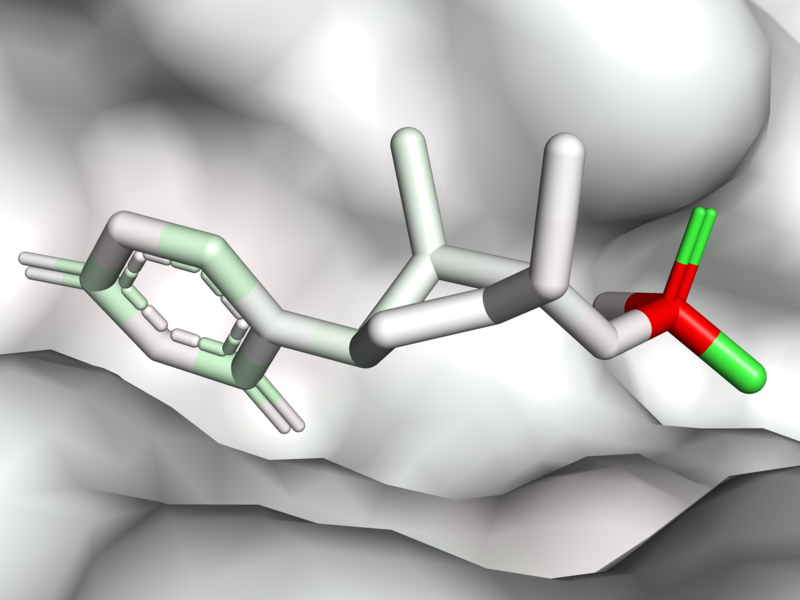}
    \caption{CLRP}
    \end{subfigure}
    \hfill
    \begin{subfigure}[t]{.32\linewidth}
    \centering
    \includegraphics[width=\linewidth]{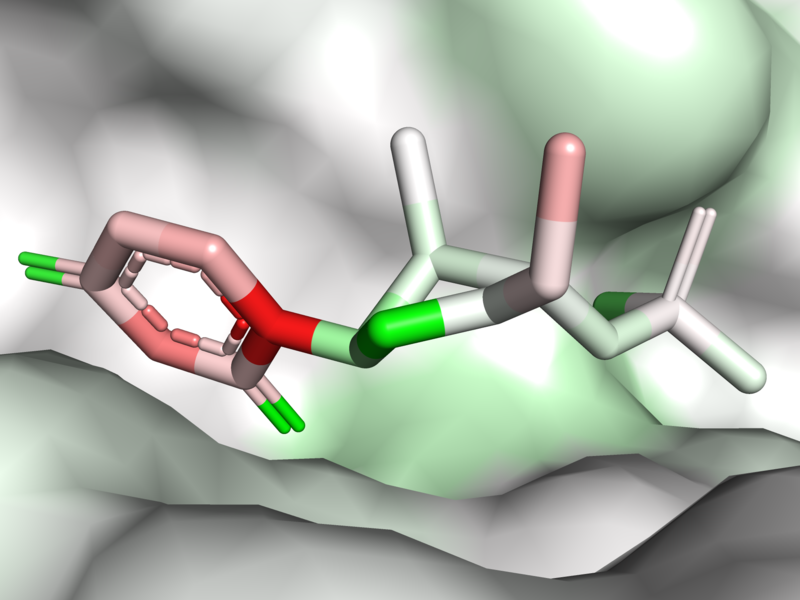}
    \caption{Masking}
    \end{subfigure}
\end{subfigure}

\begin{subfigure}[t]{\linewidth}
    \caption*{Pose Prediction Score = 0.9828}
    \begin{subfigure}[t]{.32\linewidth}
    \centering
    \includegraphics[width=\linewidth]{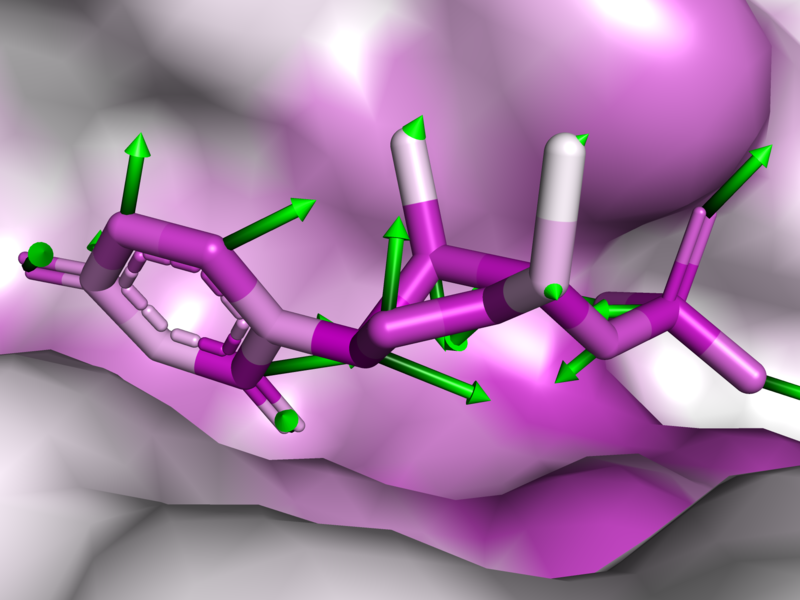}
    \caption{Gradient}
    \end{subfigure}%
    \hfill
    \begin{subfigure}[t]{.32\linewidth}
    \centering
    \includegraphics[width=\linewidth]{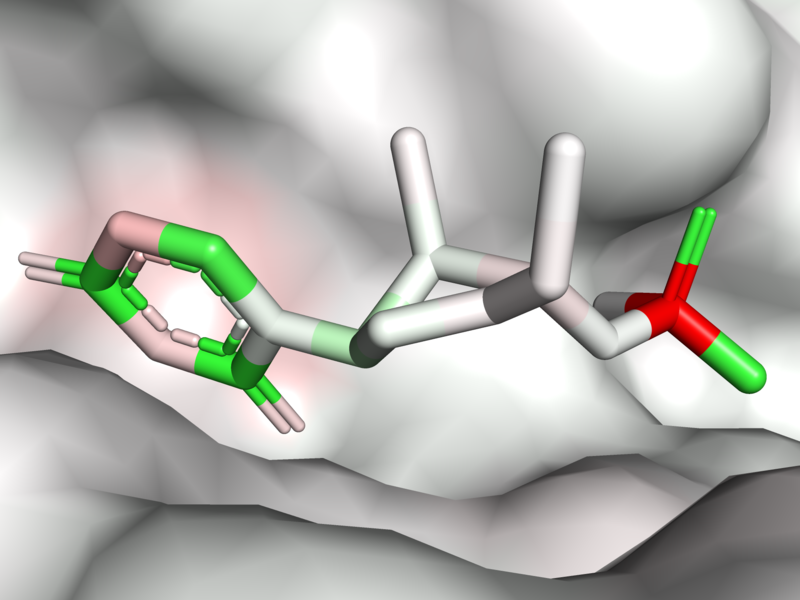}
    \caption{CLRP}
    \end{subfigure}
    \hfill
    \begin{subfigure}[t]{.32\linewidth}
    \centering
    \includegraphics[width=\linewidth]{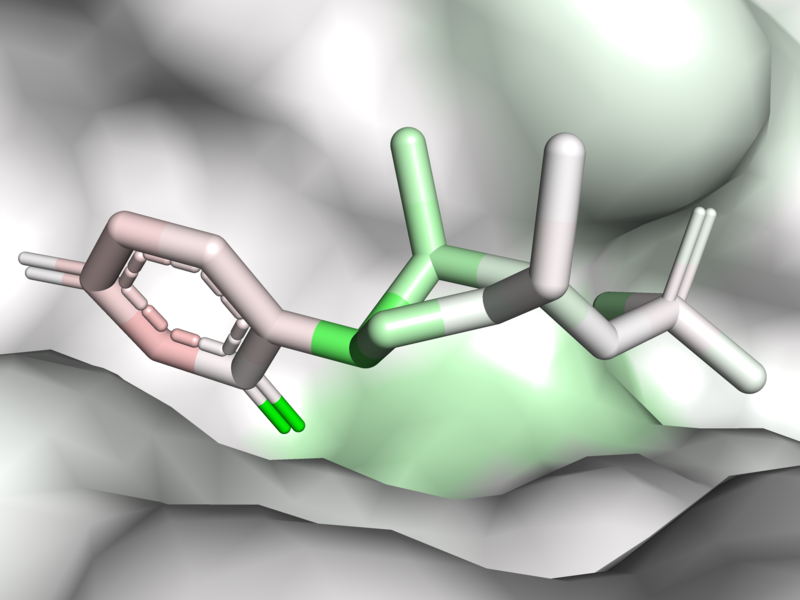}
    \caption{Masking}
    \end{subfigure}
\end{subfigure}

\caption{PDB 1w4o. The atom scores calculated with the CLRP and masking algorithms are shown as a red-white-green gradient, with green corresponding to a more favorable score.  As the gradient norms are unsigned, they are visualized as a purple gradient.}
\label{fig:1w4o}
\end{figure}

\begin{figure}[tbp]
\centering

\begin{subfigure}[t]{\linewidth}
    \caption*{Affinity Prediction Score = 5.9512}
    \begin{subfigure}[t]{.32\linewidth}
    \centering
    \includegraphics[width=\linewidth]{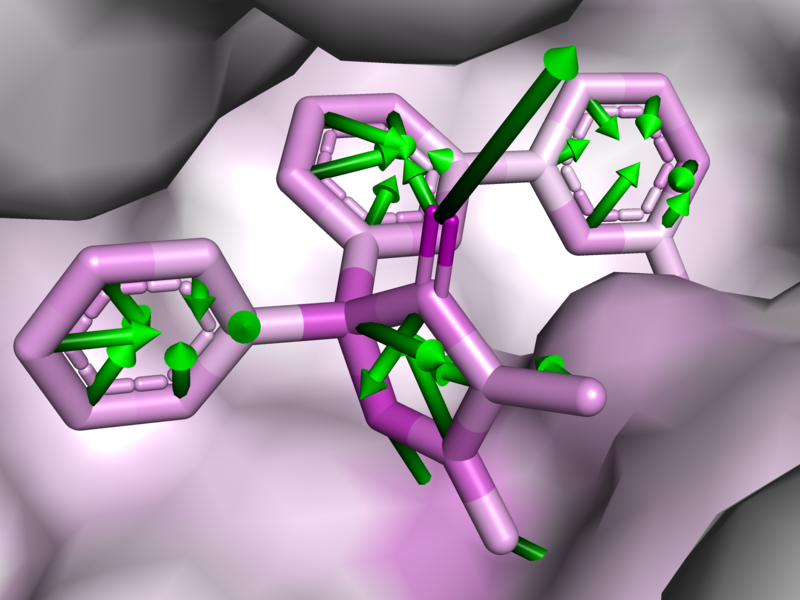}
    \caption{Gradient}
    \end{subfigure}%
    \hfill
    \begin{subfigure}[t]{.32\linewidth}
    \centering
    \includegraphics[width=\linewidth]{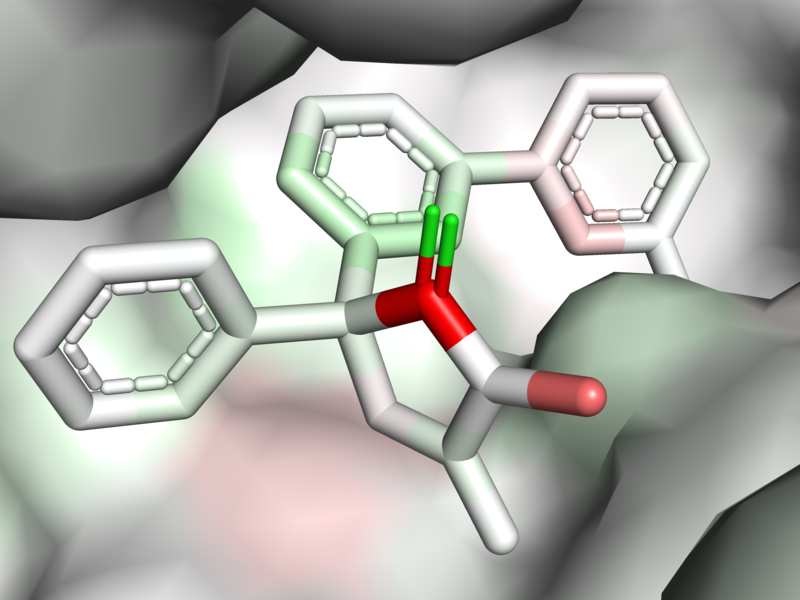}
    \caption{CLRP}
    \end{subfigure}
    \hfill
    \begin{subfigure}[t]{.32\linewidth}
    \centering
    \includegraphics[width=\linewidth]{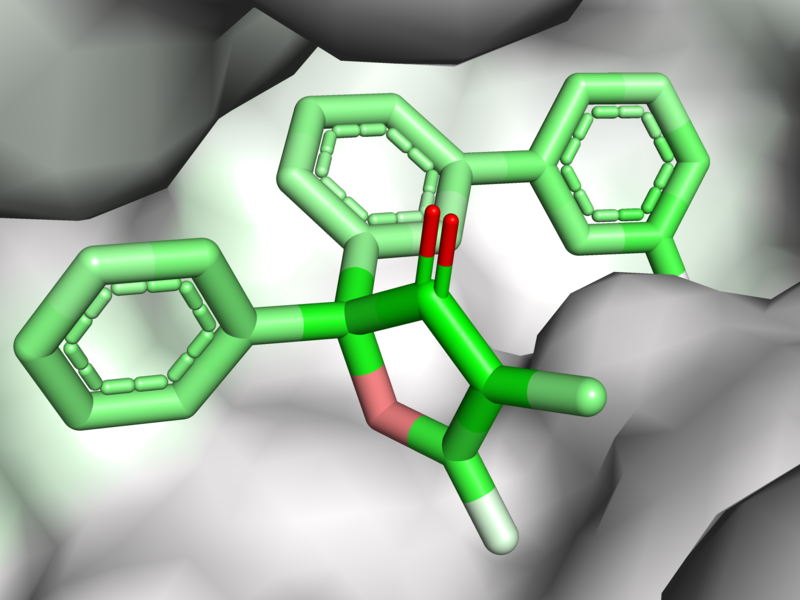}
    \caption{Masking}
    \end{subfigure}
\end{subfigure}

\begin{subfigure}[t]{\linewidth}
    \caption*{Pose Score = 0.8941}
    \begin{subfigure}[t]{.32\linewidth}
    \centering
    \includegraphics[width=\linewidth]{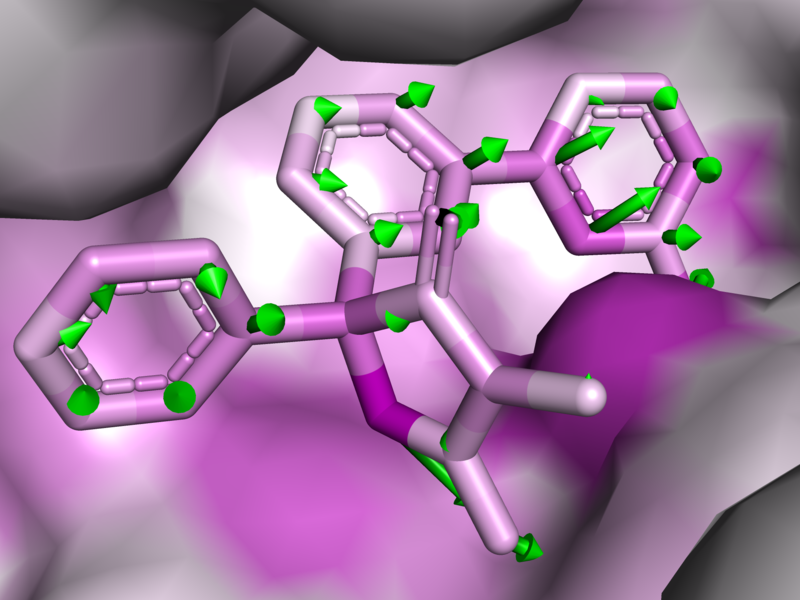}
    \caption{Gradient}
    \end{subfigure}%
    \hfill
    \begin{subfigure}[t]{.32\linewidth}
    \centering
    \includegraphics[width=\linewidth]{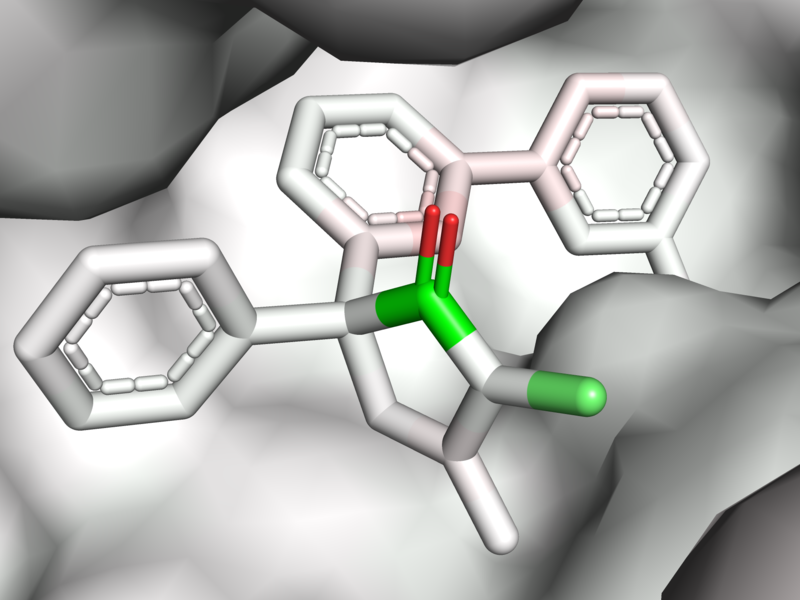}
    \caption{CLRP}
    \end{subfigure}
    \hfill
    \begin{subfigure}[t]{.32\linewidth}
    \centering
    \includegraphics[width=\linewidth]{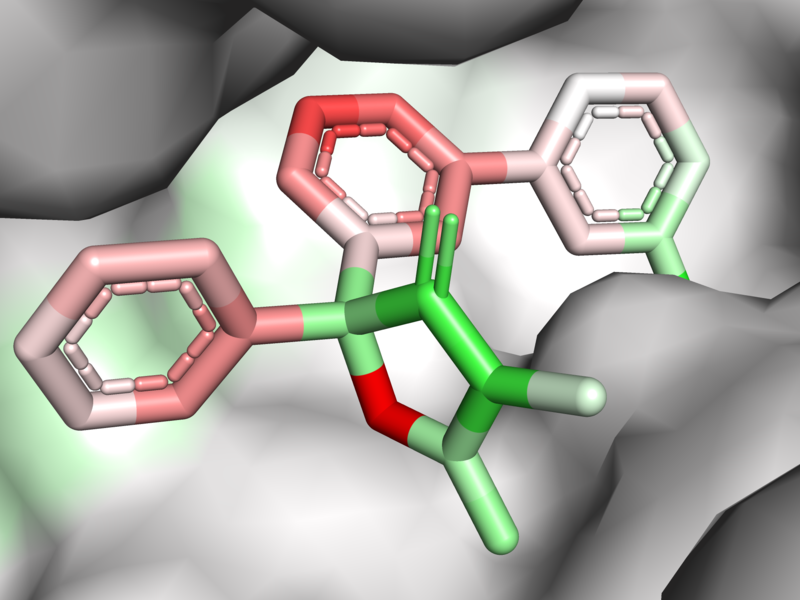}
    \caption{Masking}
    \end{subfigure}
\end{subfigure}
\caption{PDB 4djv. The atom scores calculated with the CLRP and masking algorithms are shown as a red-white-green gradient, with green corresponding to a more favorable score.  As the gradient norms are unsigned, they are visualized as a purple gradient.}
    \label{fig:4djv}
\end{figure}

\subsection{Atomistic Visualizations}

Three examples of complexes visualized with masking, gradient, and LRP methods
are shown in Figures~\ref{fig:1o0h}-~\ref{fig:1w4o}.  PyMOL session files are available at \url{http://bits.csb.pitt.edu/files/vizpaper_complexes.zip}. Examples were chosen to represent a range of scores and to have partially exposed binding sites to enable clear visualization.
Poses are maintained across the figures for ease of comparison.
Color gradients are normalized separately for each visualization method of each protein-ligand complex.  Negative (unfavorable) values are shown as red and positive (favorable) as green.  As scalar gradients are unsigned, they are shown as purple.
The crystal poses are scored after removing the water from the crystal structure.
Figure~\ref{fig:blank_examples} shows the three evaluated complexes colored by atom type. 

Figure~\ref{fig:1o0h} shows a complex, ribonuclease A with 5'-ADP, that has a low score in both affinity
prediction and pose scoring. In the shown pose, the left side of the molecule consists of phosphate groups that make a network of hydrogen bond interactions with the protein, the middle ribose group is largely solvent exposed, and the aromatic  adenine group on the right forms a network of hydrogen bonds and stacks with a histidine.
Pose scoring and affinity prediction show similar visualizations in this case. The hydrogen bonding atoms are favored, but the adenine is mostly disfavored.  The gradients show that the network wants to move this aromatic group away from the histidine, indicating that the network likely has not learned to correctly value aromatic interactions, which may explain why this micromolar compound was scored so poorly.
As the pose is low scoring, there is less relevance of the CLRP visualization to display.  Interestingly, LRP focuses on the central ribose (unfavored) which interacts with a hydrophobic valine (favored) and does not highlight the hydrogen bonding or aromatic interactions.  This may be consistent with the behavior of LRP in image recognition, where it is observed that LRP tends to highlight decision boundaries.  For example, LRP will highlight the outline of an object rather than the entirety of the object. In this case, the network may find the interaction between the hydrophilic ribose and the hydrophobic valine to be the determining feature of the complex.

Figure~\ref{fig:1w4o} shows a complex, ribonuclease A bound to a nonnatural nucleotide that contains a furanose in place of a ribose, with a relatively low affinity prediction
score and a very high pose score. Although the phosphate group is in the same place, the rest of the nucleotide extends in the opposite direction compared to Figure~\ref{fig:1o0h}. 
This pose is scored highly with much of the relevance coming from the phosphate and uracil groups. T45, which interacts with uracil and whose mutant, T45G, is known to reduce binding affinity \cite{FEBS:FEBS4511}, is one of the interacting residues highlighted as important by the affinity masking visualization.  Masking also highlights the importance of a number of the hydrogen bonding atoms, although, interestingly and contrary to CLRP, not the phosphate oxygens.

Figure~\ref{fig:4djv} shows a complex, BACE bound to an inhibitor, with a middling affinity prediction score
and a good pose score. The gradient visualizations show interesting patterns.
The arrows on the aromatic groups in the affinity prediction visualization point to the ring center, perhaps indicating the network would prefer a smaller functional group in these locations.
The arrows on the pose scoring visualization indicate a preference for a slight translation, which matches with the masking visualization, which disfavors many of the aromatic ring atoms.
Most of the visualizations strongly indicate the solvent exposed carbonyl to be important, although with sometimes conflicting interpretations of its desirability.  These inconsistencies may be an artifact of decomposing the score into individual atom contributions, as, particularly with CLRP, the carbon and oxygen of the carbonyl counter-balance each other.

\subsection{Additivity Analysis}

\begin{figure}[tbp]
\centering
\begin{subfigure}[t]{.45\linewidth}
\includegraphics[width=\linewidth]{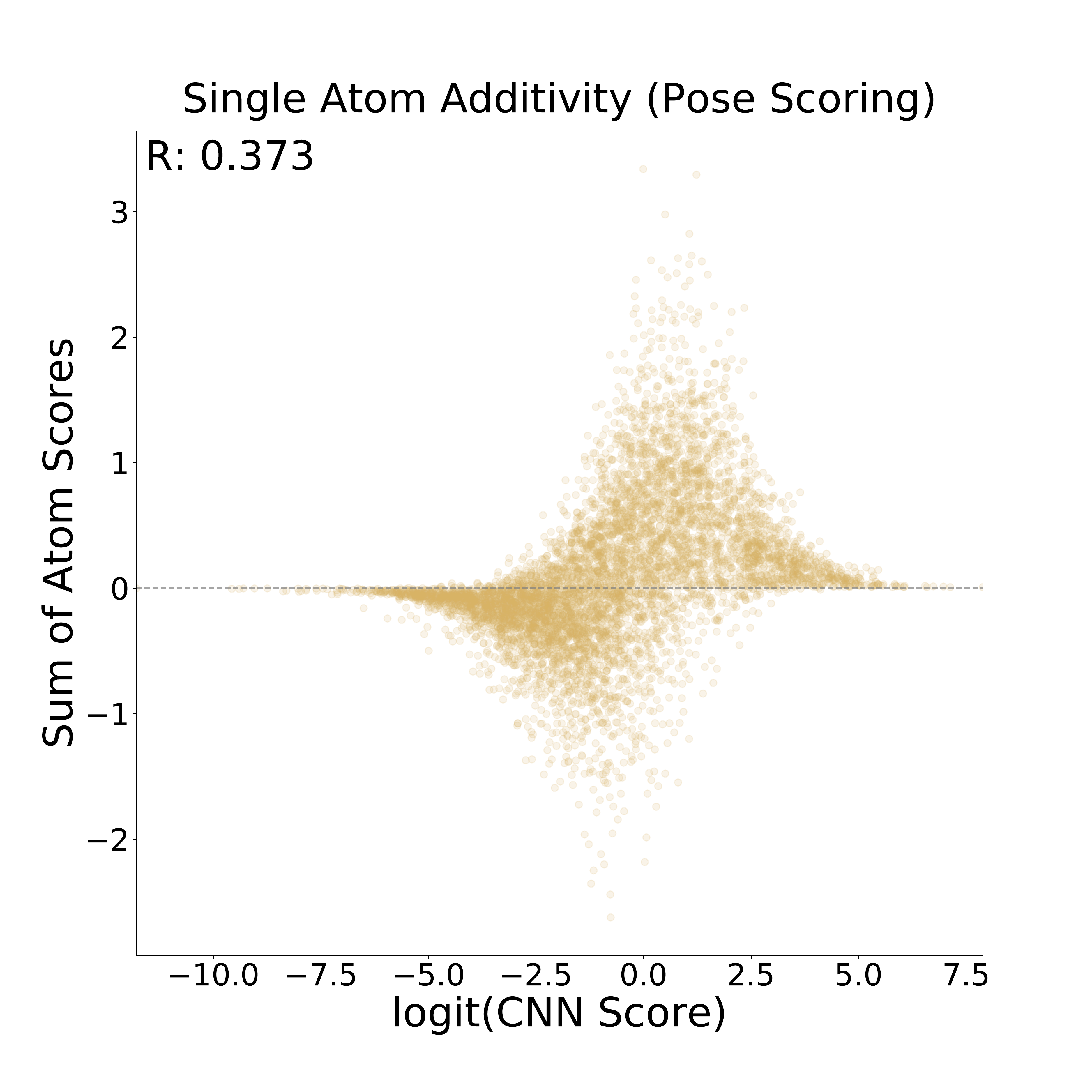}
\caption{\label{posesingle}}
\end{subfigure}%
\hfill
\begin{subfigure}[t]{.45\linewidth}
\includegraphics[width=\linewidth]{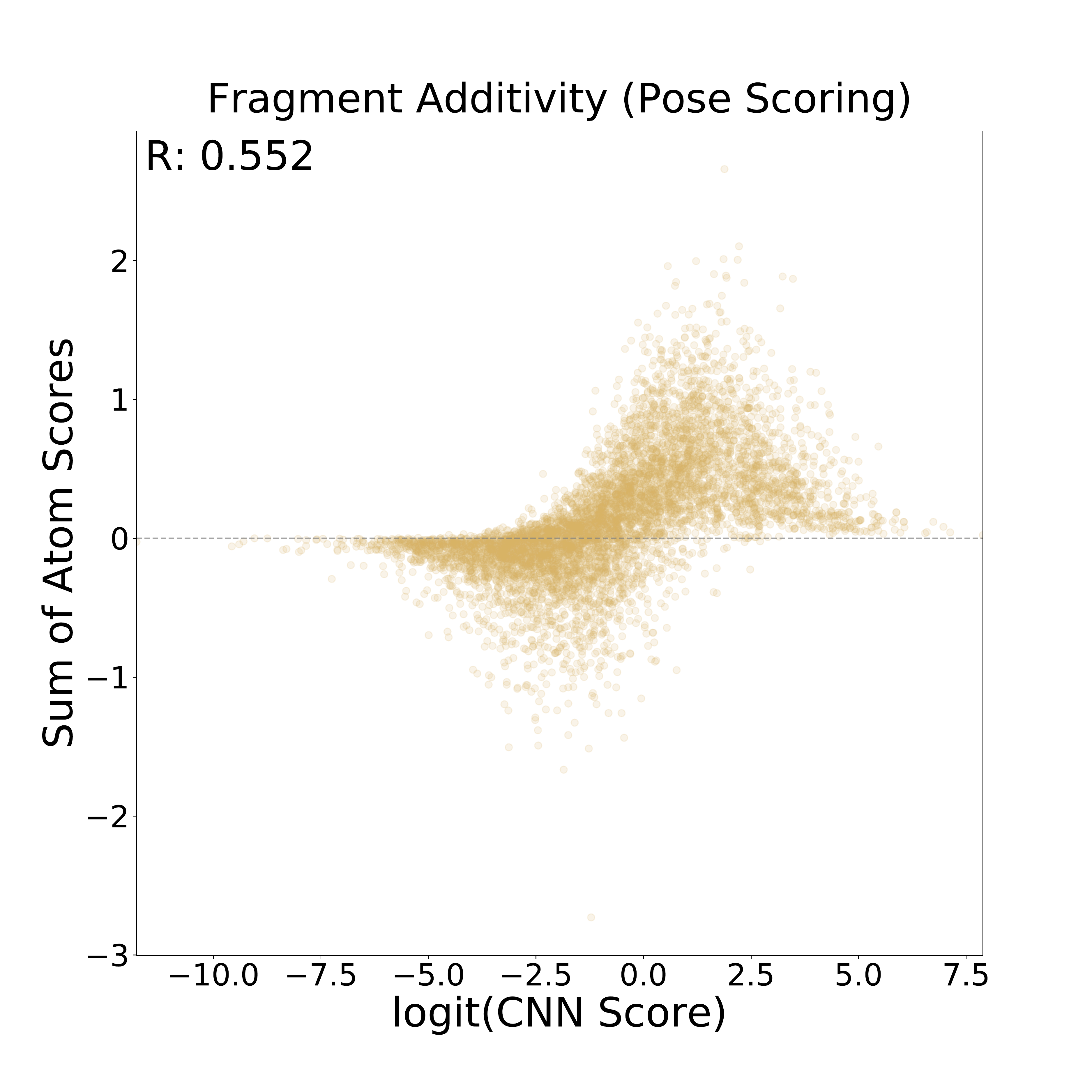}
\caption{\label{posefrag}}
\end{subfigure}

\begin{subfigure}[t]{.45\linewidth}
\includegraphics[width=\linewidth]{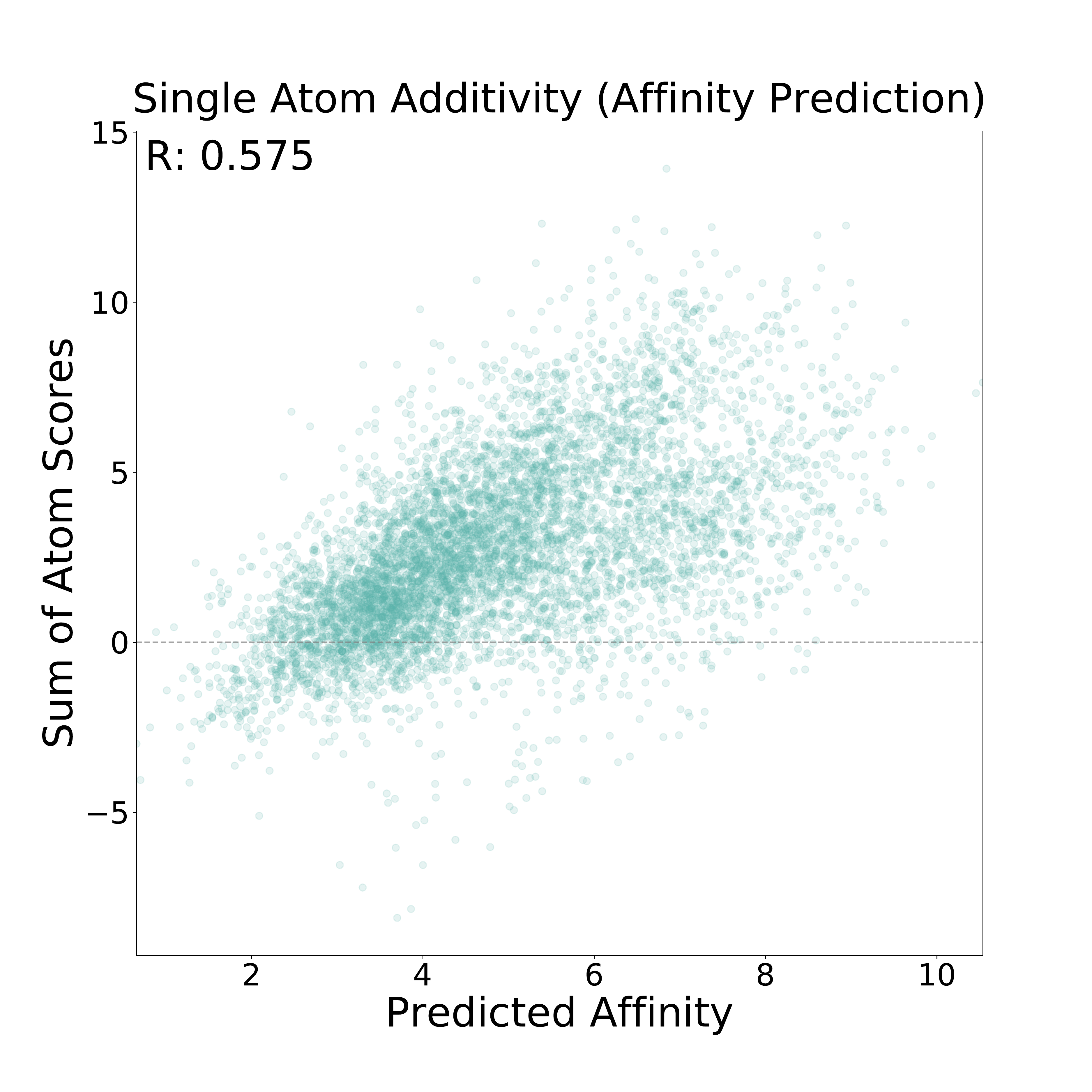}
\caption{\label{affsingle}}
\end{subfigure}%
\hfill
\begin{subfigure}[t]{.45\linewidth}
\centering
\includegraphics[width=\linewidth]{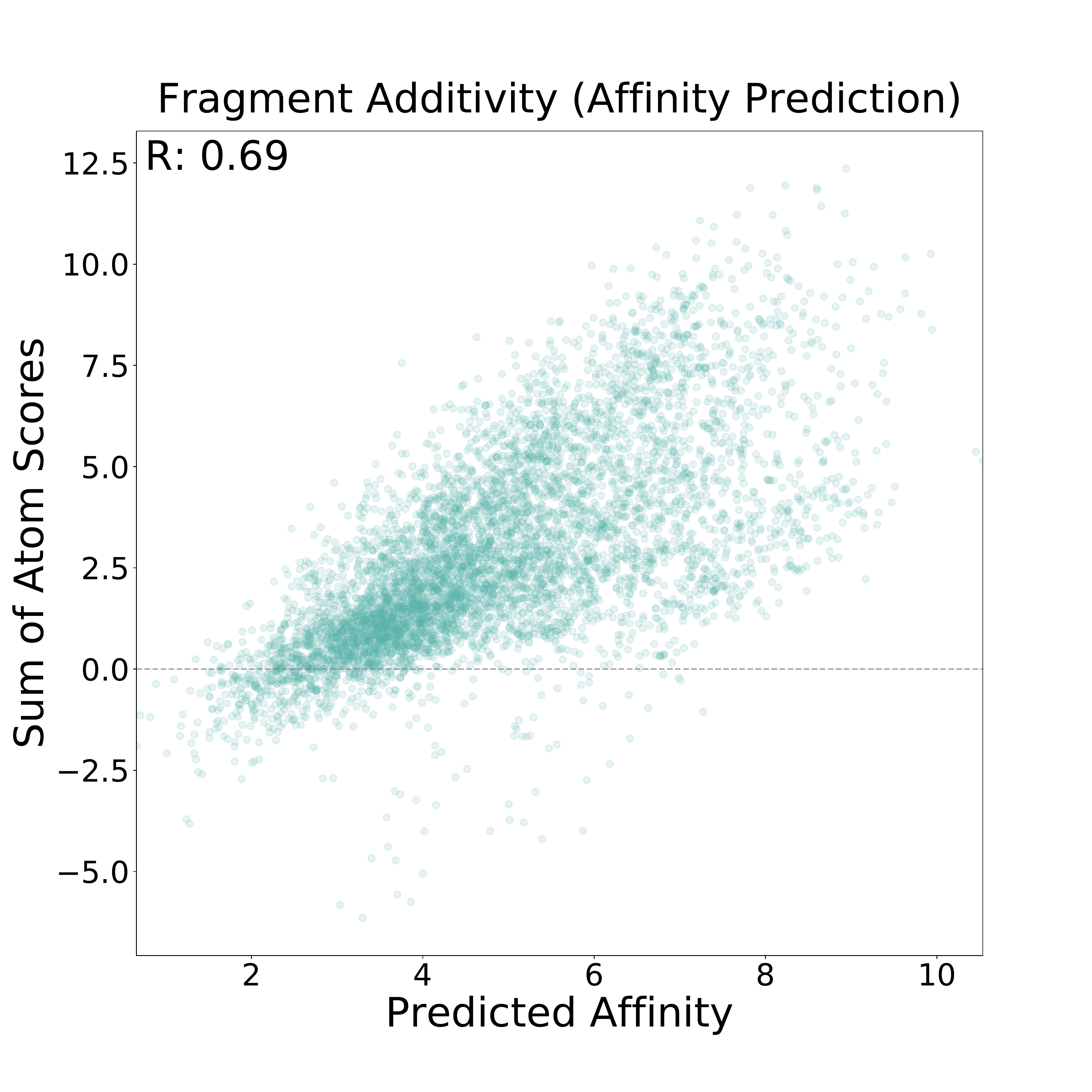}
\caption{\label{afffrag}}
\end{subfigure}
\caption{Additivity analysis. The sum of the individual atomic scores are compared to the overall score for 
\subref{posesingle} pose scoring with individual atom masking,
\subref{posefrag} pose scoring with fragment masking,
\subref{affsingle} affinity scoring with individual atom masking, and
\subref{afffrag} affinity scoring with fragment masking.
}
\label{fig:additivity}
\end{figure}

Additivity analysis evaluates the extent that the individual atomic scores constructed during masking visualization sum to the total score.  This provides a means of assessing whether or not extended atomic relationships that can't be easily decomposed into individual atomic contributions are being considered by the network.  Each atom score calculated by masking, either through single atom removal or fragment removal, is summed to a single score sum per complex. These score sums are plotted relative to the full molecular score for the complex in
Figure~\ref{fig:additivity}. A linear relationship implies that the score can be decomposed into individual atomic contributions without loss of accuracy. The analysis was carried out structures docked to the CSAR\cite{csar2010,csar2013} data set.  A non-linear relationship may indicate the network is learning more complex features of the input.

Pose scoring and affinity prediction exhibit different additivity relationships due to differences in the last layers (the convolutional layers are shared between the two).
Pose scoring imposes a softmax layer that flattens predictions between zero and one with a sigmoid function. This results in a
clustering of pose scores close to each end of the allowed range, corresponding to high and low confidence predictions. As a result,
extremely high-scoring or extremely low-scoring complexes are often unaffected by removing single atoms resulting in near-zero sums.  The effect is less pronounced with fragment removal, as this evaluates more significant modifications to the molecule.  However, for high-confidence poses, even large changes to the molecule do not reduce the perceived quality of the pose of the remaining structure. For poses with an intermediate confidence, there is a weak correlation between score sums and scores.  Affinity prediction exhibits a higher correlation with score sums, which improves when fragment masking is used.  This indicates that the network is using information about larger grouping of atoms to arrive at its predictions.

\begin{figure}[tbp]
\centering
\begin{subfigure}[t]{\linewidth}
\begin{subfigure}[t]{.45\linewidth}
\centering
\includegraphics[width=\linewidth]{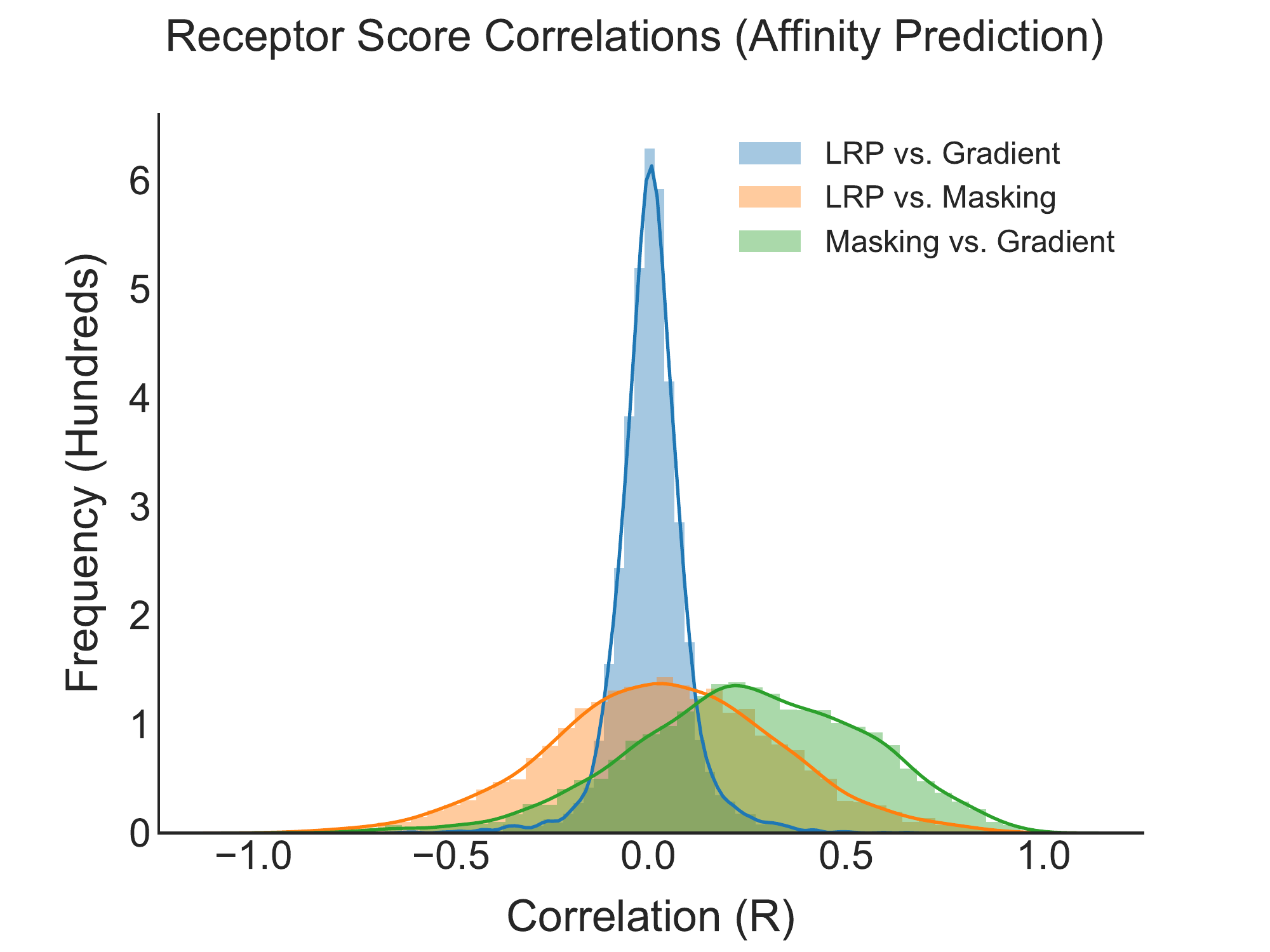}
\caption{\label{affrec}}
\end{subfigure}%
\hfill
\begin{subfigure}[t]{.45\linewidth}
\centering
\includegraphics[width=\linewidth]{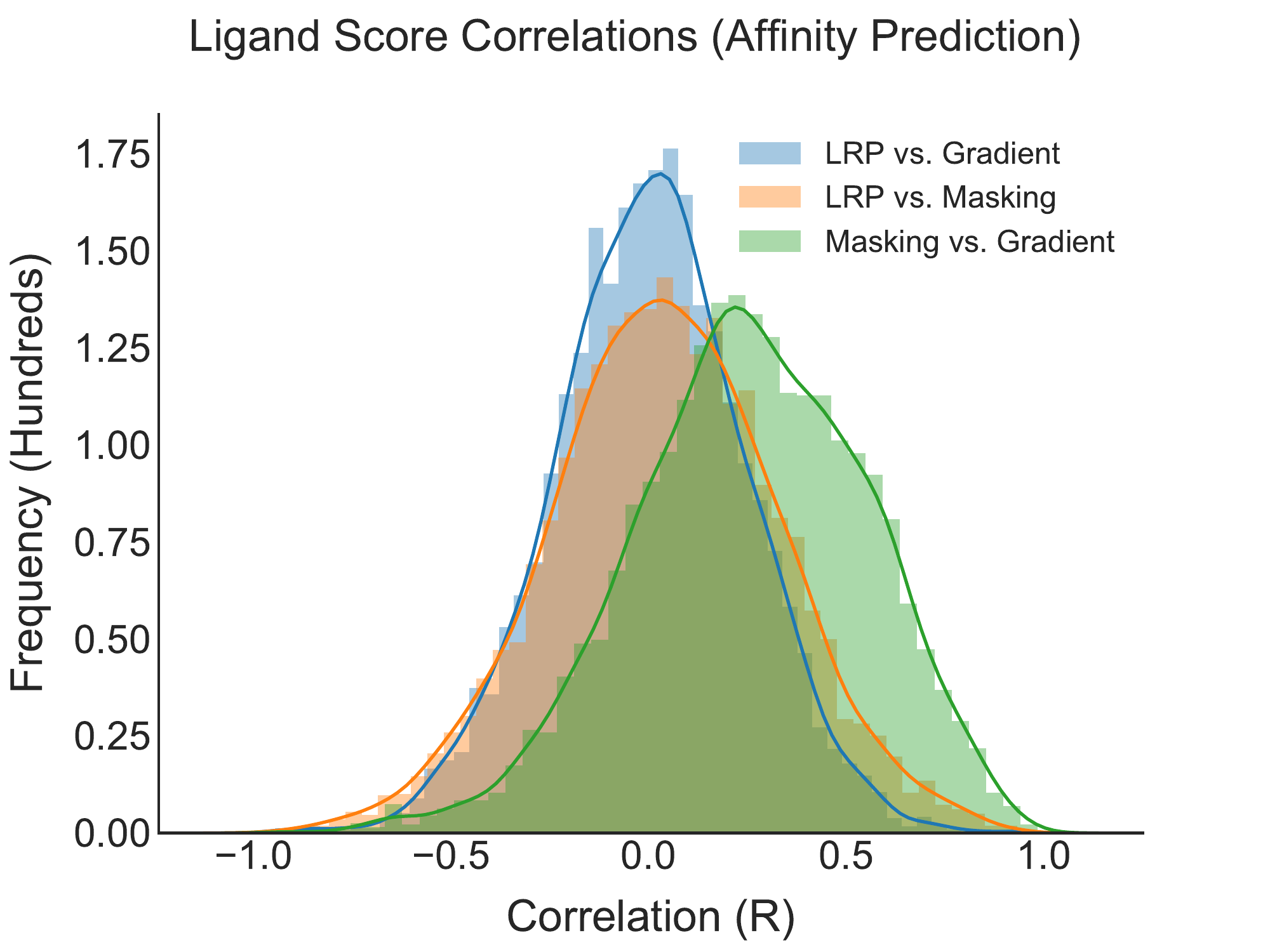}
\caption{\label{afflig}}
\end{subfigure}
\end{subfigure}

\begin{subfigure}[t]{\linewidth}
\begin{subfigure}[t]{.45\linewidth}
\includegraphics[width=\linewidth]{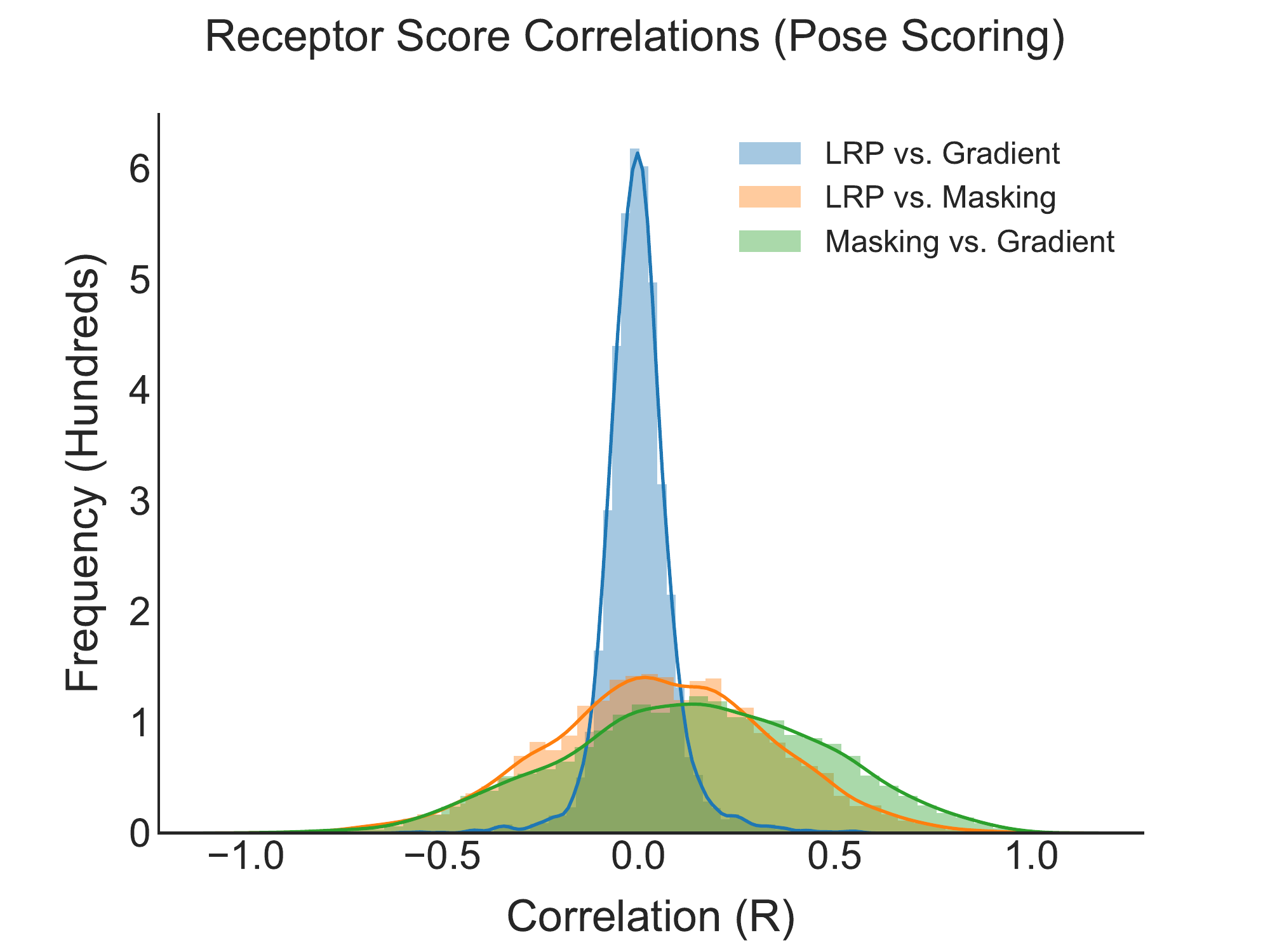}
\caption{\label{poserec}}
\end{subfigure}%
\hfill
\begin{subfigure}[t]{.45\linewidth}
\centering
\includegraphics[width=\linewidth]{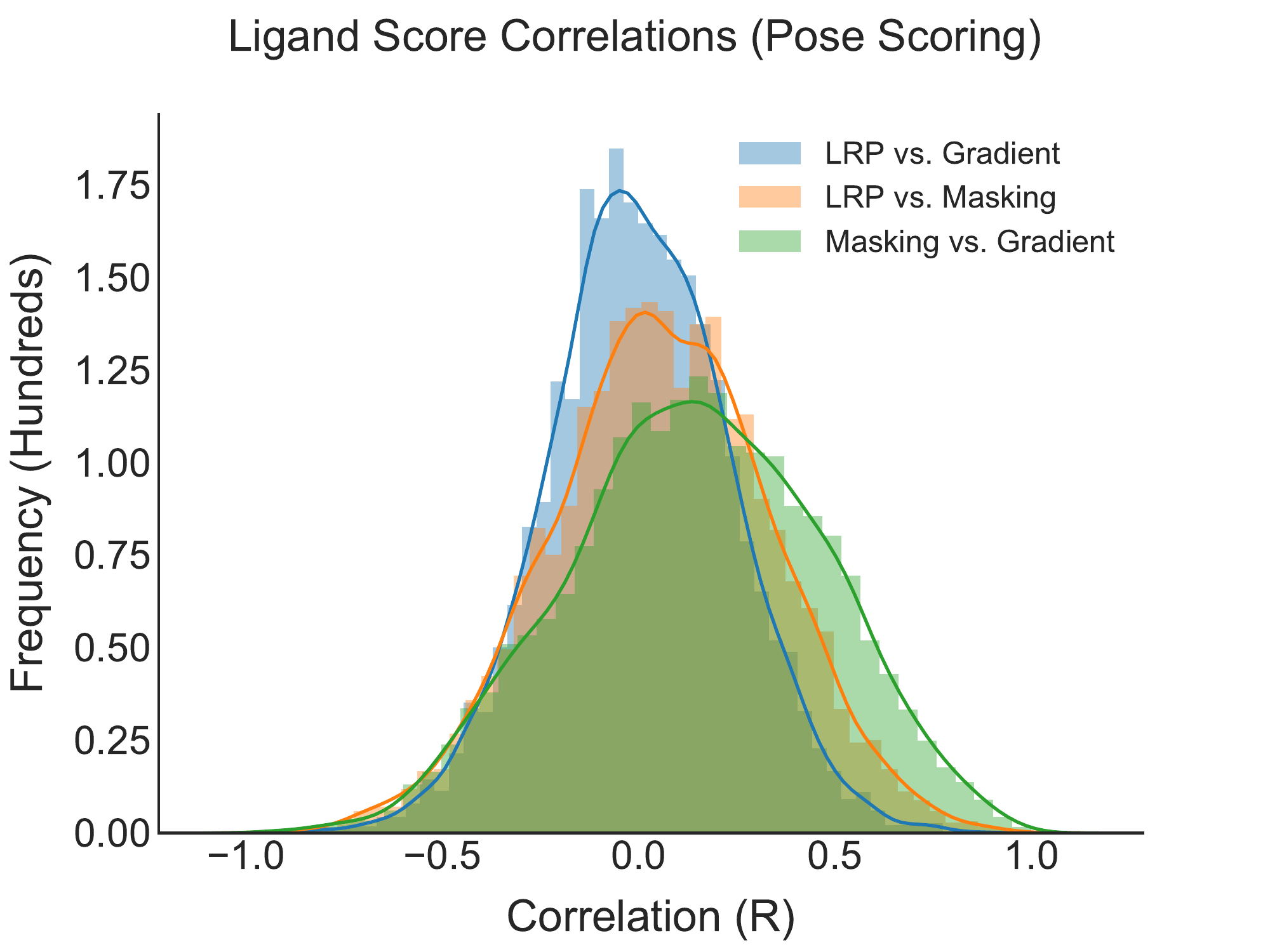}
\caption{\label{poselig}}
\end{subfigure}
\end{subfigure}
\caption{Histograms of atomic score correlations between different scoring methods. For each docked complex, the scores assigned by two visualization algorithms were compared and the correlation computed.  The distribution of scores across all the complexes is shown.
}
    \label{fig:method_correlations}
\end{figure}

\subsection{Visualization Method Comparison}

In order to compare the scores produced by each scoring method, correlations
were generated between per-atom scores in each complex for each method and are shown
in Figure~\ref{fig:method_correlations}.  In most cases, the correlations are centered around zero with a normal distribution.  However, there does appear to be some agreement between the gradient and CLRP methods, which both address the effect of changing a structure.  The general lack of correlation between methods suggests that each method presents a different interpretation of the neural network score and all three methods may provide useful insight.

\begin{figure}[tbp]
\begin{subfigure}[t]{.45\linewidth}
\centering
\includegraphics[width=\linewidth]{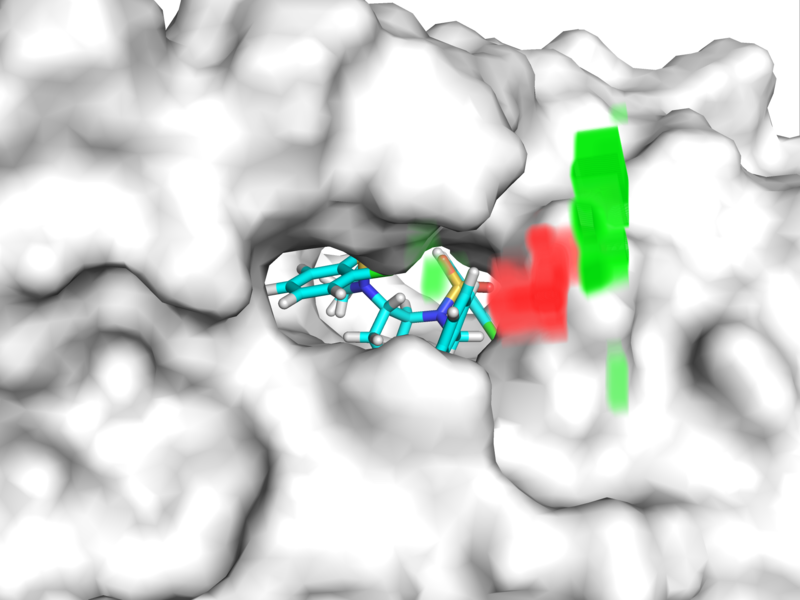}
\caption{\label{posezero}Pose}
\end{subfigure}%
\hfill
\begin{subfigure}[t]{.45\linewidth}
\centering
\includegraphics[width=\linewidth]{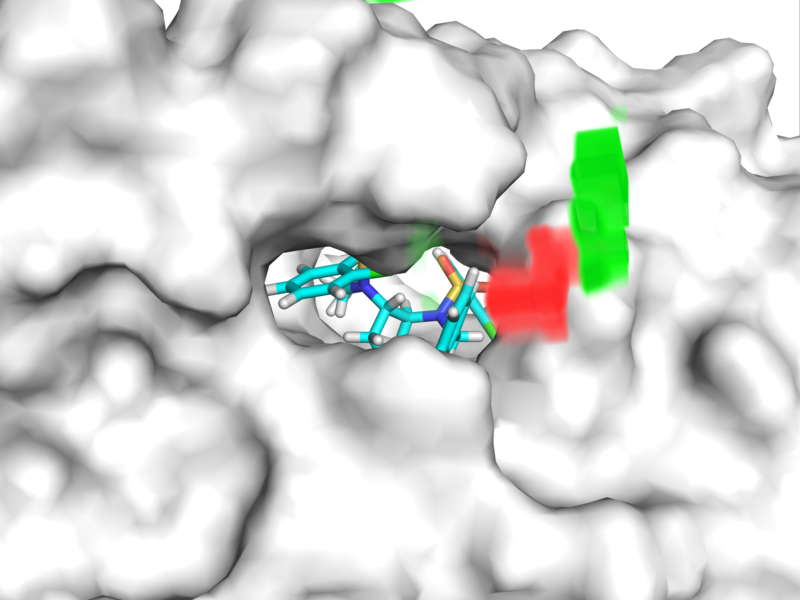}
\caption{\label{affzero}Affinity}
\end{subfigure}
\caption{Visualization of relevance generated by empty space (implicit solvent) for \subref{posezero} pose scoring and \subref{affzero} affinity scoring for PDB 2qnq. Red
    volumes are negative relevance, and green volumes are positive relevance. In this example, both the pose and affinity prediction networks generally agree on their interpretation of space outside the pocket. This complex has a predicted affinity of 6.47599 and a pose score of 0.988475.}
\label{fig:empty1}
\end{figure}

\begin{figure}[tbp]
\begin{subfigure}[t]{.45\linewidth}
\centering
\includegraphics[width=\linewidth]{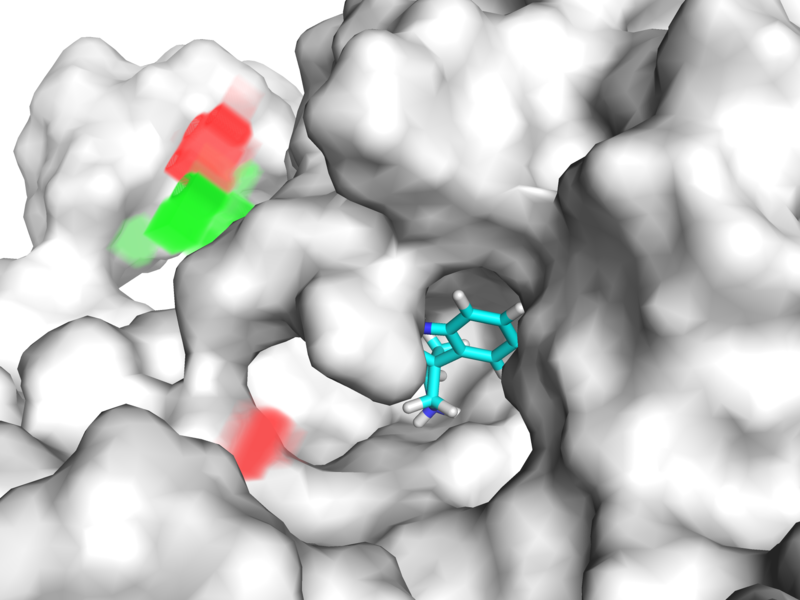}
\caption{\label{posezero2}Pose}
\end{subfigure}%
\hfill
\begin{subfigure}[t]{.45\linewidth}
\centering
\includegraphics[width=\linewidth]{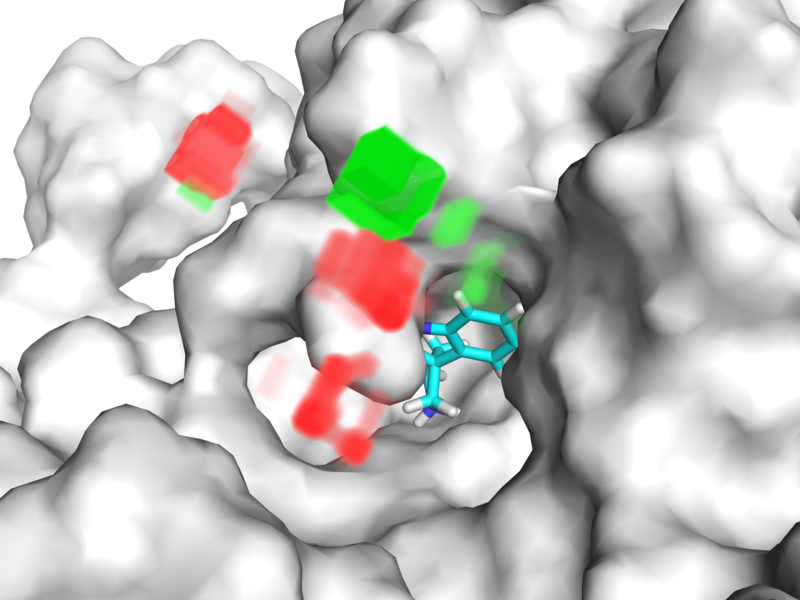}
\caption{\label{affzero2}Affinity}
\end{subfigure}
\caption{Visualization of relevance generated by empty space (implicit solvent) for \subref{posezero2} pose scoring and \subref{affzero2} affinity scoring for PDB 3udh. Red
    volumes are negative relevance, and green volumes are positive relevance.  The pose and affinity networks diverge, with the front opening of the pocket having high relevance to affinity prediction, but not for pose scoring. This complex has a predicted affinity of 5.02264 and a pose score of 0.881191. }
\label{fig:empty2}
\end{figure}

\subsection{Empty Space Visualization}

Any atom-based visualization method has the limitation that it will not visualize the contribution of the implicit solvent to the final score. However, during the computation of the CLRP algorithm we record the relevance that would be absorbed by ``dead nodes'' with a pre-activation value of zero.  We find that, on average, more than 99\% of dead nodes are in the first layer. This suggests that they are a consequence of implicit solvent features directly encoded in the input, rather than higher order features that are found later in the network. By mapping the relevance values of dead nodes of this first layer onto the input grid, we can see where in the input dead nodes are occurring, and how the network treats them. We find these nodes are almost exclusive the result of empty space, areas with no atom density corresponding to the implicit solvent.

Two examples of empty space visualization are shown in Figures~\ref{fig:empty1} and ~\ref{fig:empty2}.  In Figure~\ref{fig:empty1}, an HIV protease inhibitor, both the affinity and pose score view the empty space immediately outside the binding site negatively.  This implies that a complex would be scored more favorably if either the ligand or receptor were to fill this space.  It is worth noting that some HIV protease inhibitors due extend into this area (e.g. PDB 5ivr). Farther from the binding site, the empty space contributes positive relevance, indicating leaving this area solvent exposed is preferred. Figure~\ref{fig:empty2} shows a BACE1 inhibitor.  In this case the relevance of the empty areas is significantly different between affinity and pose scoring and suggests that extending the ligand out of the pocket would improve affinity, but would not necessarily increase the confidence in the correctness of the current pose.

\section{Discussion}

Each visualization method provides a different insight into how the CNN is scoring different inputs. Masking is the most intuitive approach, and arguably produces the most understandable results. It approaches the CNN from an external perspective, as it operates outside the network itself by manipulating the input. However, masking incurs a significant computational cost. In contrast, gradient and CLRP visualizations can be generated with a single forwards/backwards pass through the network. Masking has the advantage of being analogous to experimental methods in determining important interactions in protein-ligand complexes, and thus may be the best suited for interpreting the results produced by the CNN. 

Gradient visualizations are a useful tool in determining what the network ``wants'' for a particular input to be better. Because the gradients can be generated on a per-atom basis, they provide a more specific breakdown of how the network is interpreting the input. Gradients have the benefit of providing directional information and can be used to refine poses \cite{ragoza2017ligand}.

CLRP has the advantage of preserving relevance all the way to the input, which is not the case with the epsilon or alpha-beta stabilization of LRP. It also does not have parameters that can produce different results, as is the case with both epsilon and alpha-beta stabilization. It does have a drawback, in that the ``dead'' nodes are completely ignored. This can be partially rectified by dead node analysis, which provides information about how the network considers empty space.

On the whole, the pose scoring and affinity prediction networks evaluate complexes similarly. Figures~\ref{fig:1o0h}-~\ref{fig:1w4o} compare the two networks, and the masking and gradient visualizations correlate. The CLRP visualization are more contrasted, and often are more difficult to interpret, possibly because it focuses on highlighting the decision boundary of the network . This could mean that CLRP has limited usefulness in this domain, or that it is showing relationships that are beyond our current ability to interpret. Either way, it has the most utility in analyzing the effect of empty space in the input. 

Visualization serves two main purposes: to guide medicinal chemistry optimization and to inform the construction and training of the network.  All three visualization techniques clearly highlight import atoms of the ligand, which can then be targeted for modification by medicinal chemistry.  Combined with convolutional filter visualization, they also provide different ways of interpreting the network and improving performance.  For example,  Figure~\ref{fig:1o0h} illustrates that the network puts little value on an aromatic interaction.  Assuming this observation is replicated in additional incorrectly scored molecules, this insight could be leveraged to retrain the network with a training set enriched in aromatic interactions.  We also note that the visualizations described here should generalize to other molecular deep learning methods.  A complete implementation, \texttt{gninavis}, is available under an open-source license as part of the gnina project (\url{http://github.com/gnina}.

\section*{Acknowledgements}
We thank Jocelyn Sunseri, Justin Spiriti, and Paul Francoeur for their feedback during the preparation of the manuscript.   This work is supported by R01GM108340 from
the National Institute of General Medical Sciences and by a GPU donation from the NVIDIA corporation.




\bibliographystyle{model1-num-names}
\bibliography{references.bib}







\end{document}